\documentclass[10pt,journal,compsoc]{IEEEtran}
\ifCLASSOPTIONcompsoc
  \usepackage[nocompress]{cite}
\else
  \usepackage{cite}
\fi
\ifCLASSINFOpdf
\else
\fi
\usepackage{amsmath,amssymb,amsfonts,mathtools}

\usepackage{breqn}

\def\P{\mathbb{P}} 
\def\E{\mathbb{E}} 

\def\R{\mathbb{R}}

\def\KL{\text{KL}}
\def\H{\text{H}}

\renewcommand{\vec}[1]{\boldsymbol{\mathbf{#1}}}
\usepackage{color, colortbl}
\definecolor{Gray}{gray}{0.97}

\usepackage{amsthm}
\newtheorem{theorem}{Theorem}

\newtheorem{proposition}{Proposition}

\usepackage{microtype}
\usepackage{footnote}
\makesavenoteenv{tabular} \makesavenoteenv{table}
\usepackage{graphicx}
\usepackage{booktabs} 
\usepackage{enumerate}
\usepackage{caption}
\usepackage{subcaption}

\usepackage{comment}
\usepackage{color}
\usepackage{mathtools}
\usepackage{systeme}

\usepackage{multirow}
\usepackage{hyperref}
\usepackage{breqn}
\usepackage[export]{adjustbox}
\usepackage{enumitem}
\usepackage{textcomp}
\begin{document}
%
\title{Adversarial Robustness via Fisher-Rao Regularization}
%
%
%
%

\author{
Marine~Picot, Francisco~Messina, Malik~Boudiaf, Fabrice~Labeau,~\IEEEmembership{Senior Member,~IEEE,}\\
~Ismail~Ben~Ayed, and~Pablo~Piantanida,~\IEEEmembership{Senior Member,~IEEE}

\IEEEcompsocitemizethanks{\IEEEcompsocthanksitem  M. Picot, F. Messina, and F. Labeau  are with the Department of Electrical and Computer Engineering, McGill University, QC, Canada.\protect\\ Email:\{marine.picot,francisco.messina\}@mail.mcgill.ca 
\IEEEcompsocthanksitem Malik~Boudiaf and Ismail~Ben~Ayed are with ETS Montreal
\protect\\
Email:\{malik.boudiaf.1, Ismail.BenAyed\}@etsmtl.net
\IEEEcompsocthanksitem M. Picot is with Laboratoire des Signaux et Syst\`emes (L2S), Universit\'e Paris-Saclay CNRS CentraleSupélec, Gif-sur-Yvette, France. Email:marine.picot@centralesupelec.fr
\IEEEcompsocthanksitem P. Piantanida is with the International Laboratory on Learning Systems (ILLS), McGill - ETS - Mila - CNRS - Universit\'e Paris Saclay - CentraleSupelec. Email:pablo.piantanida@centralesupelec.fr 
}
}

%
%

\markboth{IEEE TRANSACTIONS ON PATTERN ANALYSIS AND MACHINE INTELLIGENCE}%
{Picot \MakeLowercase{\textit{et al.}}: FIRE: Adversarial Robustness via Fisher-Rao Regularization}
%




\IEEEtitleabstractindextext{
\begin{abstract}
Adversarial robustness has become a topic of growing interest in machine learning since it was observed that neural networks tend to be brittle. We propose an information-geometric formulation of adversarial defense and introduce \textsc{Fire}, a new Fisher-Rao regularization for the categorical cross-entropy loss, which is based on the geodesic distance between {  the softmax outputs corresponding to} natural and perturbed input features. Based on the information-geometric properties of the class of softmax distributions, we derive an explicit characterization of the Fisher-Rao Distance (\textsc{FRD}) for the binary and multiclass cases, and draw some interesting properties as well as connections with standard regularization metrics. \textcolor{black}{Furthermore, we verify on a simple linear and Gaussian model, that all Pareto-optimal points in the accuracy-robustness region can be reached by \textsc{Fire} while other state-of-the-art methods fail}.
Empirically, we evaluate the performance of various classifiers trained with the proposed loss on standard datasets, showing up to { a simultaneous 1\% of improvement in terms of clean and robust performances} while reducing the training time by 20\% over the best-performing methods.
\end{abstract}

}

\maketitle

\IEEEdisplaynontitleabstractindextext

%
\IEEEpeerreviewmaketitle

\IEEEraisesectionheading{\section{Introduction}\label{Intro}}

%
%
%
%


\IEEEPARstart{D}{eep} Neural Networks (DNNs) have achieved several breakthroughs in different fields such as computer vision, speech recognition, and Natural Language Processing (NLP). Nevertheless, it is well-known that these systems are extremely sensitive to small perturbations on the inputs \cite{szegedy2013}, known as adversarial examples. Formally, an adversarial example represents a corrupted input, characterized by a bounded optimal perturbation from the original vector, designed to fool a specified neural networks' task. Adversarial examples have already proven threatful in several domains, including vision and NLP~\cite{alzantot2018generating}, hence leading to the emergence of the rich area of adversarial machine learning~\cite{vorobeychik2018}. The effectiveness of adversarial examples has been attributed to the linear regime of DNNs~\cite{goodfellow2014} and the data manifold geometrical structure itself~\cite{gilmer2018}, among other hypotheses. More recently, it has been related to the existence of valuable features for classification but meaningless for humans~\cite{ilyas2019}. 

In this paper, we focus on the so-called white-box attacks, for which the attacker has full access to the model. However, it should be noted that black-box attacks, in which the attacker can only query predictions from the model without access to further information, are also feasible \cite{papernot2016b}. The literature on adversarial machine learning is extensive and can be divided into three overlapping groups, studying the generation, detection, and defense aspects. The simplest method to generate adversarial examples is the Fast Gradient Sign Method (FGSM) \cite{goodfellow2014}, including its iterative variant called Projected Gradient Descent (PGD) \cite{madry2018}. Although widely used, PGD has a few issues that can lead to overestimating the robustness of a model. AutoAttack \cite{croce2020reliable} has been recently developed to overcome those problems, enabling an effective way to test and compare the different defensive schemes. 

A simple approach to cope with corrupted examples is to detect and discard them before classification. For instance, \cite{feinman2017}, \cite{zheng2018}, and \cite{grosse2017} present different methods to detect corrupted inputs. Although these ideas can be useful to ensure robustness to outliers (i.e., inputs with large deviations with respect to clean examples), they do not seem to be satisfactory solutions for mild adversarial perturbations. In addition, adversarial detection can generally be bypassed by sophisticated attack methods~\cite{carlini2017b}. 

Recently, several works focused on improving the robustness of neural networks by investigating various defense mechanisms. 
For instance, certified defense mechanisms addressed the need for more guarantees on the task performance beyond standard evaluation metrics \cite{li2019certified,lecuyer2019certified,cohen2019certified, croce2019provable,wong2018scaling,zhang2019towards,gowal2018effectiveness,mirman2018differentiable}. These methods aim at training classifiers whose predictions at any input feature will remain constant within a set of neighborhoods around the original input. However, these algorithms do not achieve state-of-the-art performance yet. Also, some approaches tend to rely on convex relaxations of the original problem~\cite{raghunathan2018,wong2018} since directly solving the adversarial problem is not tractable. Although these solutions are promising, it is still not possible to scale them to high-dimensional datasets. Finally, we could mention distillation, initially introduced in \cite{hinton2015}, and further studied in~\cite{papernot2016a}. The idea of distillation is to use a large DNN (the teacher) to train a smaller one (the student), which can perform with similar accuracy while utilizing a temperature parameter to reduce sensitivity to input variations. The resulting defense strategy may be efficient for some attacks but can be defeated with the standard Carlini-Wagner attack. 

In this work, we will focus on the most popular strategy for enhancing robustness, which is based on adversarial training, i.e., learning with an augmented training set containing adversarial examples~\cite{goodfellow2014}.

\subsection{Summary of contributions}
Our work investigates the problem of optimizing the trade-off between accuracy and robustness and advances state-of-the-art methods in very different ways.
\begin{itemize}[leftmargin=*]
	\item 
	We derive an explicit characterization of the Fisher-Rao Distance (\textsc{FRD}) based on the information-geometric properties of the soft-predictions of the neural classifier. That leads to closed-form 
	expressions of the \textsc{FRD} for the binary and multiclass cases (Theorems \ref{thm:binary_frd} and \ref{thm:multiclass_frd}, respectively). We further relate them to well-known regularization metrics (presented in Proposition \ref{thm:fisher_rao_bound_binary}). 
	\item 
	We propose a new formulation of adversarial defense, called FIsher-rao REgularizer (\textsc{Fire}). It consists of optimizing a regularized loss, which encourages the predictions of natural and perturbed samples to be close to each other, according to the manifold of the softmax distributions induced by the neural network. 
	Our loss in Eq. \eqref{eq:regularized_risk_fire} consists of two terms: the categorical cross-entropy, which favors natural accuracy, and a Fisher-Rao regularization term, which increases
	adversarial robustness. 
	Furthermore, we prove for a simple logistic regression and Gaussian model that all Pareto-optimal points in the accuracy-robustness region can be reached by \textsc{Fire}, while state-of-the-art methods fail (cf. Section \ref{sec:gaussian_example} and Proposition~\ref{prop-pareto}). 
	\item Experimentally, 
	on standard benchmarks, we found that FIRE provides an improvement up to roughly 2\% of robust accuracy compared to the widely used Kullback-Leibler regularizer \cite{zhang2019theoretically}. We also observed significant improvements over other state-of-the-art methods. In addition, our method typically requires, on average, less computation time (measured by the training runtime on the same GPU cluster) than state-of-the-art methods.
\end{itemize}

\vspace{-0.8em}
\subsection{Related work}

\textbf{Adversarial training.}
Adversarial training (AT)~\cite{goodfellow2014} is one of the few defenses that has not been broken so far. Indeed, different variations of this method have been proposed. It is based on an attack-defense scheme where the attacker's goal is to create perturbated inputs by maximizing a loss to fool the classifier, while the defender's goal is to classify those attacked inputs rightfully.

\textbf{Inner attack generation.}
The inner attacker design is vital to AT since it has to create meaningful attacks. One of the most popular algorithms to generate adversarial examples is Projected Gradient Descent (PGD) \cite{goodfellow2014,madry2018}, which is an iterative attack: the output at each step is the addition of the previous output and the sign of the loss gradient modulated by a fixed step size. The loss maximized in PGD is often the same loss that is minimized for the defense. 

\textbf{Robust defense loss.}
An essential choice of the defense mechanism is the robust loss used to attack and defend the network. Initially, \cite{goodfellow2014} used the adversarial cross-entropy. However, it was shown that one way to improve adversarial training is through the choice of this loss. TRADES \cite{zhang2019theoretically} introduces a robustness regularizer based on the Kullback-Leibler divergence. MART \cite{wang2019improving} uses a robustness regularizer that considers the misclassified inputs and boosted losses. 

\textbf{Additional improvement.}
Whether it is AT, TRADES or MART, they all have been improved in recent years. Those improvements can either rely on pretraining \cite{hendrycks2019using}, early stopping \cite{rice2020overfitting}, curriculum learning \cite{atzmon2019controlling}, adaptative models \cite{huang2020self}, unlabeled data to improve generalization \cite{carmon2019unlabeled,alayrac2019labels} or additional perturbations on the model weights \cite{wu2020adversarial}. 

It should be noted that the main disadvantage of adversarial training-based methods remains the required computational expenses. Nevertheless, as will be shown in Section \ref{sec:experiments}, \textsc{Fire} can significantly reduce them.

\section{Background} \label{sec:background_adv_learning}


We consider a standard supervised learning framework where $\vec{x} \in \mathcal{X} \subseteq \R^n$ denotes the input vector on the space $\mathcal{X}$ and $y \in \mathcal{Y}$ the class variable, where $\mathcal{Y} \coloneqq  \{ 1, \ldots, M \}$. The unknown data distribution is denoted by $p(\vec{x},y) = p(\vec{x}) p(y|\vec{x})$. We define a classifier to be a parametric soft-probability model of $p(y|\vec{x})$, denoted as $q_{\vec{\theta}}(y|\vec{x})$, where $\vec{\theta} \in \Theta$ are the parameters. This can be readily used to induce a hard decision: $f_{\vec{\theta}} : \mathcal{X} \to \mathcal{Y}$ with ${f_{\vec{\theta}}(\vec{x}) \coloneqq \arg \max_{y\in \mathcal{Y}}~ q_{\vec{\theta}}(y|\vec{x})}$. Adversarial examples are denoted as $\vec{x}^\prime = \vec{x} + \vec{\delta} \in \mathcal{X}$, where $\| \vec{\delta}\| \le \varepsilon$ for an arbitrary norm $\| \cdot \|$. Loss functions are denoted as $\ell(\vec{x}, \vec{x^{\prime}},y,\vec{\theta})$ and the corresponding risk functions by  $L(\vec{\theta})$. We also define the natural missclassification probability as $P_e(\vec{\theta}) \doteq \P( f_{\vec{\theta}}(\vec{X}) \ne Y)$, the adversarial missclassification probability as $P_e^{\prime}(\vec{\theta}) \doteq \P( f_{\vec{\theta}}(\vec{X}^{\prime}) \ne Y)$.



{  

\subsection{Adversarial learning} \label{sec:adv_training}

We provide some background on adversarial learning,  focusing on adversarial defense's most popular proposed loss functions. Adversarial examples are slightly modified inputs that can fool a target classifier. Concretely, Szegedy {\it et al.}~\cite{szegedy2013} define the adversarial generation problem as: 
\begin{equation}
   \vec{x}^{\prime} =\underset{\vec{x}^{\prime} \, :\, \lVert \vec{x}^{\prime} - \vec{x} \rVert_{p} < \varepsilon}{\text{arg min}}\lVert\vec{x}^{\prime} - \vec{x}\rVert\text{~~s.t.~~}f_{\theta}(\vec{x}^{\prime}) \neq y,
   \label{eq:adversarial_problem}
\end{equation}
where $y$ is the true label (supervision) associated to the sample $\vec{x}$. This formulation shows that the vulnerable points of a classifier are the ones close to its decision boundaries. Since this problem is difficult to tackle, it is commonly relaxed as follows~\cite{carlini2017}: 
\begin{equation}
   \vec{x}^{\prime} = \underset{\vec{x}^{\prime} \,:\,\lVert\vec{x}^{\prime} - \vec{x}\rVert_{p}<\varepsilon
   }{\text{arg max}} \ell(\vec{x},\vec{x}^{\prime},y, \vec{\theta}).
   \label{eq:relaxed_adv_problem}
\end{equation}
Once adversarial examples are obtained, they can be used to learn a robust classifier as discussed next.

The adversarial problem has been presented in \cite{madry2018} as follows: 
\begin{equation}
\min_{\vec{\theta}} ~ \E_{p(\vec{x},y)} \left[ \max_{\vec{x}^{\prime} : \|\vec{x}^{\prime} - \vec{x} \|_{p}\leq \varepsilon } \; \ell(\vec{x}, \vec{x}^{\prime},y,\vec{\theta}) \right], \end{equation}
where $\varepsilon$ denotes the maximal distortion allowed in the adversarial examples according to the $l_p$-norm. Since the exact solution to the above inner max problem is generally intractable, a relaxation is proposed by generating an adversarial example using an iterative algorithm such as PGD. }


\subsubsection{Adversarial Cross-Entropy (ACE)} If we take the loss to be the Cross-Entropy (CE), i.e., { $\ell(\vec{x}, \vec{x}^{\prime},y, \vec{\theta}) = - \log q_{\vec{\theta}}(y|\vec{x}^{\prime})$}, we obtain the ACE risk:
\begin{equation} \label{eq:ace_loss} L_{\text{ACE}}(\vec{\theta}) \doteq \E_{p(\vec{x},y)} \left[ \max_{\vec{x}^{\prime} : \|\vec{x}^{\prime} - \vec{x} \|_{p}\leq \varepsilon}  - \log q_{\vec{\theta}}(y|\vec{x}^{\prime}) \right]. \end{equation}


\subsubsection{TRADES} Later \cite{zhang2019theoretically} defined a new risk based on a trade-off between natural and adversarial performances, controlled through an hyperparameter $\lambda$. The resulting risk is the addition of the natural cross-entropy and the Kullback-Leibler (KL) divergence between natural and adversarial probability distributions: 
{ 
\begin{align} L_{\textrm{TRADES}}(\vec{\theta})  \doteq \E_{p(\vec{x},y)} \Big[ &\max_{\vec{x}^{\prime} : \|\vec{x}^{\prime} - \vec{x} \|_{p}\leq \varepsilon} - \log q_{\vec{\theta}}(y|\vec{x})  \nonumber \\ &+ \lambda ~
 \text{KL} (q_{\vec{\theta}}(\cdot|\vec{x})\|q_{\vec{\theta}}(\cdot|\vec{x}^{\prime})) \Big] \label{eq:trades-loss}, \end{align}}
where
\begin{equation}
\text{KL} (q_{\vec{\theta}}(\cdot|\vec{x})\|q_{\vec{\theta}}(\cdot|\vec{x}^{\prime})) \doteq \E_{q_{\vec{\theta}}(y|\vec{x})}\left[ \log \frac{ q_{\vec{\theta}}(y|\vec{x})}{q_{\vec{\theta}}(y|\vec{x}^{\prime})} \right].    
\end{equation}

\section{Adversarial Robustness with Fisher-Rao Regularization}

\subsection{Information Geometry and Statistical Manifold} \label{sec:motivation}
\begin{figure}
	\begin{center}
		\includegraphics[width=0.3\textwidth]{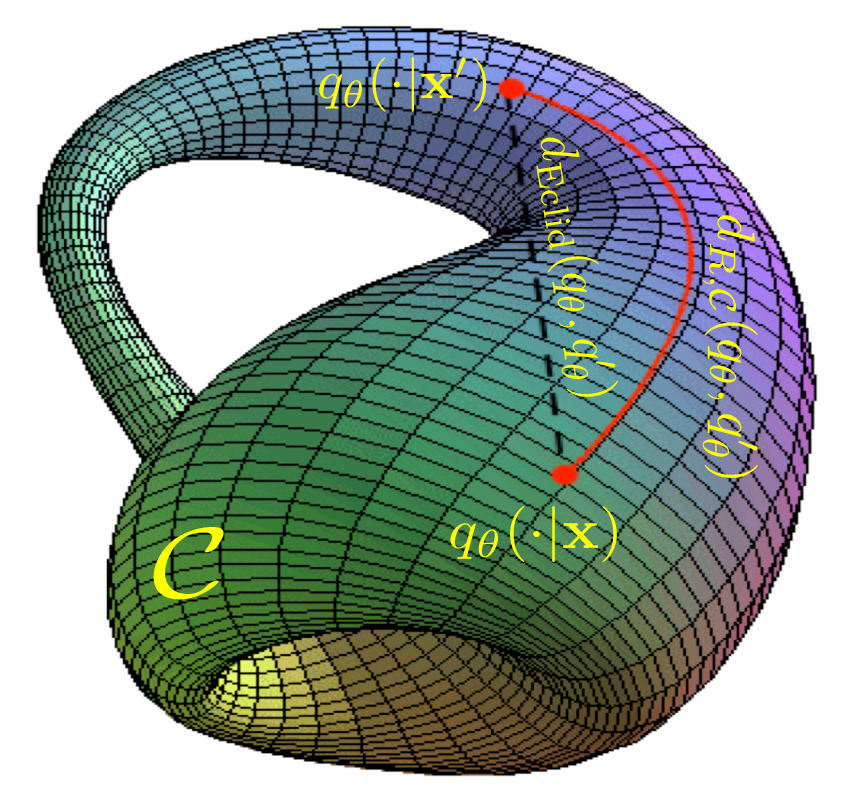}
	\end{center}
	\caption{Illustration of \textsc{FRD} between two distributions $q_{\vec{\theta}}= q_{\vec{\theta}}(\cdot|\vec{x})$ and $q_{\vec{\theta}}^{\prime} = q_{\vec{\theta}}(\cdot|\vec{x}^{\prime})$ over the statistical manifold $\mathcal{C}$.}
	\label{fig:frd_illustration}
\end{figure}

Statistics on manifolds and information geometry are two different ways in which differential geometry meets statistics. A statistical manifold can be defined as a parameterized family of  probability distributions  (or density functions) of interest. It is worth to mention that the concept of  statistics on manifolds is very different from manifold learning which  is a branch of machine learning where the goal is to learn a latent manifold from valued data. In this paper, we are interested in the statistical manifold obtained when fixing the parameters $\vec{\theta}$ of a DNN and changing its feature input. We consider the following statistical manifold: \mbox{$\mathcal{C} \doteq \big\{ q_{\vec{\theta}}(\cdot|\vec{x}) : \vec{x} \in \mathcal{X} \big\}$}. In particular, the focus is on changes in a neighborhood of a particular sample in an adversarial manner (i.e., considering a worst-case perturbation).  { Please notice that the statistical manifold is different from the loss landscape. The loss landscape is defined as the changes of the risk function with respect to changes in the model parameters (i.e., $L(\vec{\theta})$ vs $\vec{\theta}$), while the statistical manifold  refers to the changes of the soft-probabilities of the classifier with respect to changes in the input (i.e., $q_{\vec{\theta}}(y|\vec{x})$ vs $\vec{x}$).} In order to understand the effect of a perturbation on the input, we first need to be able to capture the distance over the statistical manifold between different probability distributions, i.e., between two different feature inputs. That is precisely what the \textsc{FRD} computes, as illustrated by the red curve in Fig. \ref{fig:frd_illustration}. It is worth to mention that \textsc{FRD} can be very much different from the euclidean distance since the later does not depend on the shape of the manifold. For a formal mathematical definition of \textsc{FRD}  and a short review of basic concepts in information geometry, we refer the reader to Section \ref{app:review_frd}. 

\begin{figure}[t!]
	\centering
	\begin{subfigure}[t]{0.5\columnwidth}
		\centering
		\includegraphics[height=1in]{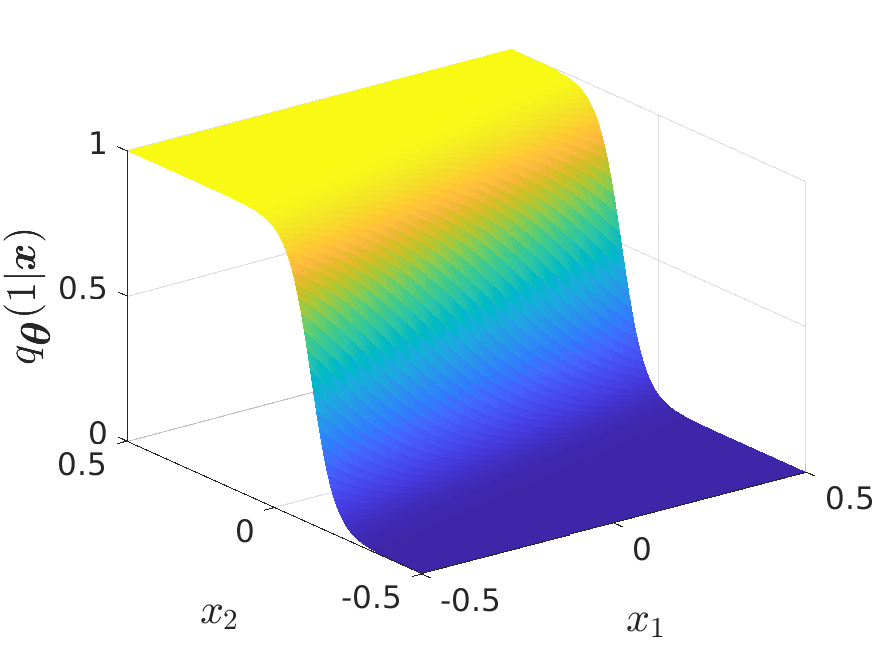}
		\caption{$\vec{\theta} = [-6.4290, 25.7487]^{\intercal}$.}
		\label{fig:statistical_manifold_nat}
	\end{subfigure}%
	~ 
	\begin{subfigure}[t]{0.5\columnwidth}
		\centering
		\includegraphics[height=1.3in]{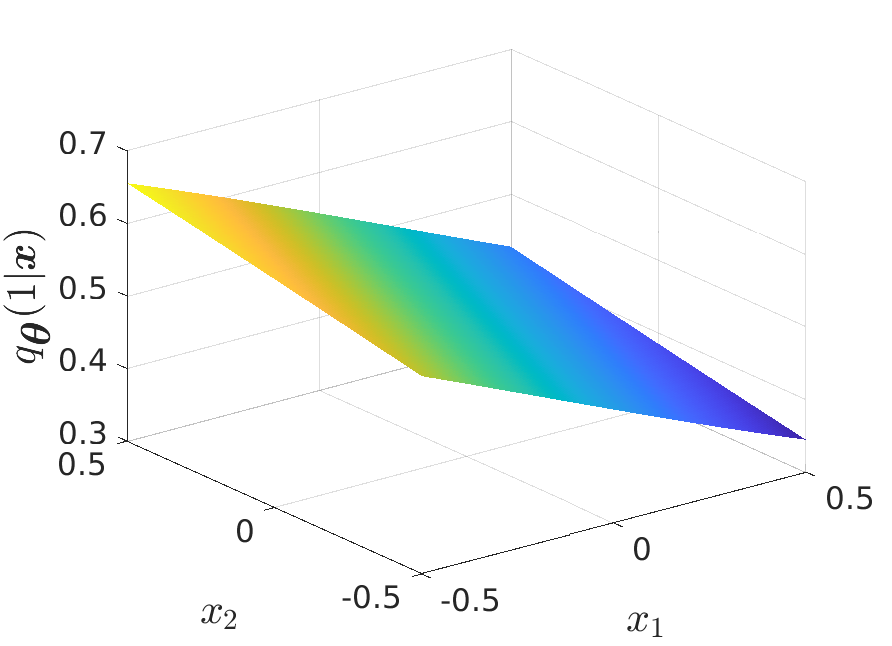}
		\caption{$\vec{\theta} = [-0.9364, 0.3509]^{\intercal}$.}
		\label{fig:statistical_manifold_adv}
	\end{subfigure}
	\caption{Visualization of statistical manifold $\mathcal{C}$ defined by the model $q_{\vec{\theta}}(y|\vec{x}) = 1/[1 + \exp(- y \, \vec{\theta}^{\intercal} \vec{x})]$ with different values of $\vec{\theta}$: (a) Parameters minimizing the natural misclassification error probability $P_e$, (b) Parameters minimizing the adversarial misclassification error probability $P_e^{\prime}$.}
	\label{fig:statistical_manifolds}
\end{figure}

{  As discussed in Section \ref{sec:background_adv_learning}, the robustness of a classifier is related to the distance between natural examples and the decision boundaries (i.e., points $\vec{x}$ such that $q_{\vec{\theta}}(y|\vec{x}) \approx q_{\vec{\theta}}(y'|\vec{x})$ for $y \ne y'$). In fact, if a natural example is far from the decision boundaries, a norm-constrained attack will clearly fail (in this case, the optimization problem \eqref{eq:adversarial_problem} will be infeasible). Since the decision boundaries are given by the soft-probabilities $q_{\vec{\theta}}(y|\vec{x})$, this can be equivalently studied by analyzing the shape of the statistical manifold \mbox{$\mathcal{C}$} (which should not be confused with the loss landscape). In fact, if $q_{\vec{\theta}}(y|\vec{x})$ is \textcolor{black}{relatively flat (i.e., does not change much)} with respect to perturbations of $\vec{x}$ around $\vec{x}_0$, it is clear that adversarial perturbations will not modify the classifier decision at this point. In contrast, if $q_{\vec{\theta}}(y|\vec{x})$ changes sharply with perturbations of $\vec{x}$ around $\vec{x}_0$, an adversarial can easily leverage this vulnerability to fool the classifier.} \textcolor{black}{This notion of robustness is related to the Lipschitz constant of the network, as discussed in various works (e.g., \cite{pmlr-v70-cisse17a}).} To illustrate these ideas clearly, let us consider the logistic regression model $q_{\vec{\theta}}(y|\vec{x}) = 1/[1 + \exp(- y \, \vec{\theta}^{\intercal} \vec{x})]$, where $n = 2$ {  and $\mathcal{Y} = \{-1,1\}$}, as a simple example. One way to visualize the statistical manifold $\mathcal{C}$ is to plot $q_{\vec{\theta}}(1|\vec{x})$ as a function of $\vec{x}$ {  (since $q_{\vec{\theta}}(-1|\vec{x}) = 1 - q_{\vec{\theta}}(1|\vec{x}) $, this completely characterizes the manifold)}. This is shown in Fig. \ref{fig:statistical_manifold_nat} for the value of $\vec{\theta}$ which minimizes the natural missclassification probability $P_e$ under a conditional Gaussian model for the input $\vec{x}$ (see Section \ref{sec:gaussian_example} for details). As can be seen, the manifold is quite sharp around a particular region of $\mathcal{X}$. This region corresponds to the neighborhood of the points for which $\vec{\theta}^\intercal \vec{x} \approx 0$ as $\vec{x}$ is perturbed in the direction of $\vec{\theta}$. Therefore, we can say that this model is clearly non-robust, as its output can be significantly changed by small perturbations on the input. Consider now the same model but with the values of $\vec{\theta}$ obtained by minimizing the adversarial missclassification probability $P_e^\prime$. As can be seen in Fig. \ref{fig:statistical_manifold_adv}, the statistical manifold is much flatter than in Fig. \ref{fig:statistical_manifold_nat}, which means that the model is {  less sensitive to} adversarial perturbations on the input. {  Therefore, it is more robust.}


\begin{figure}[t!]
	\centering
	\begin{subfigure}[t]{0.5\columnwidth}
		\centering
		\includegraphics[height=1.3in]{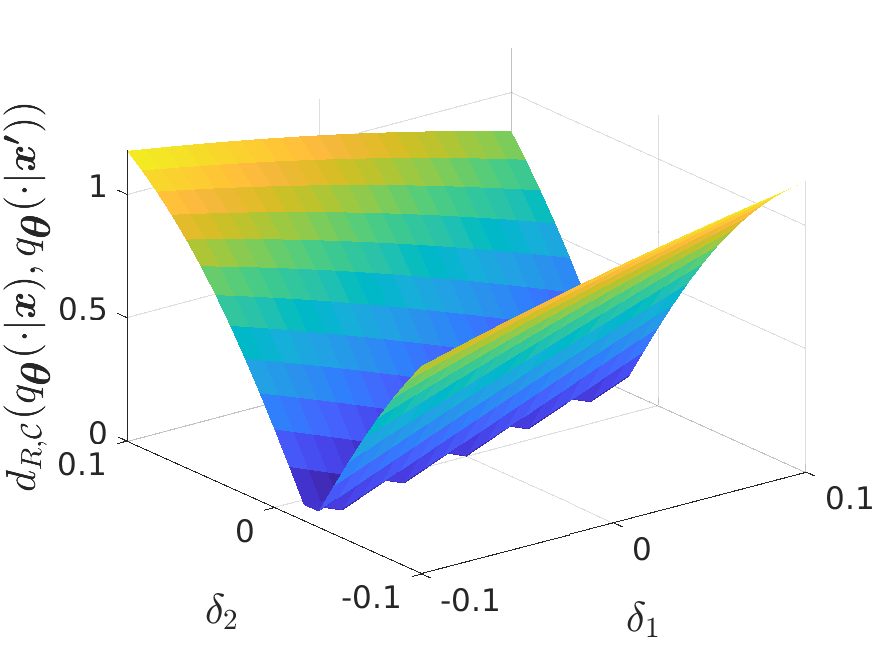}
		\caption{$\vec{\theta} = [-6.4290, 25.7487]^{\intercal}$.}
		\label{fig:frd_nat}
	\end{subfigure}%
	~ 
	\begin{subfigure}[t]{0.5\columnwidth}
		\centering
		\includegraphics[height=1.3in]{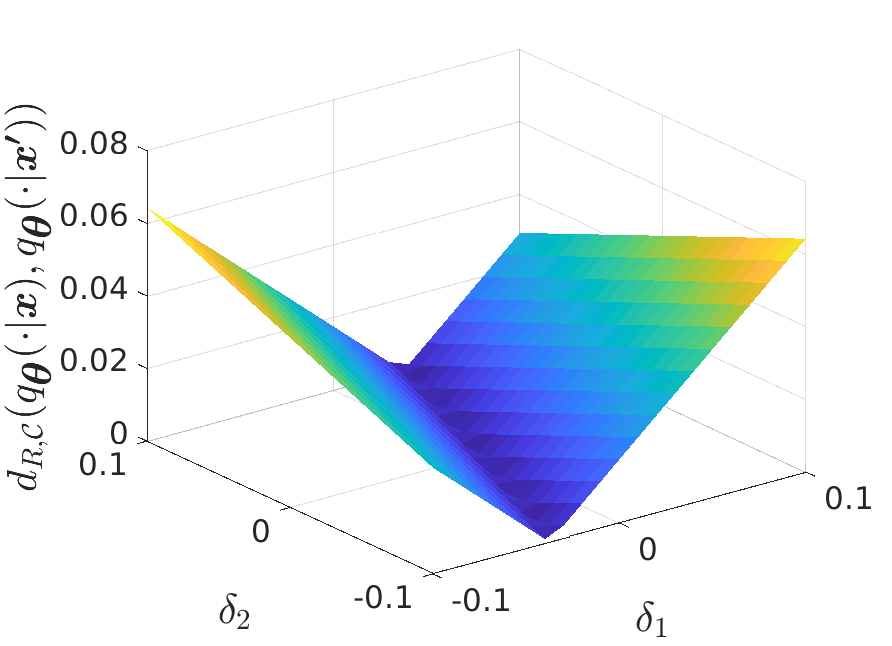}
		\caption{$\vec{\theta} = [-0.9364, 0.3509]^{\intercal}$.}
		\label{fig:frd_adv}
	\end{subfigure}

	\caption{\textsc{FRD} between the distributions $q_{\vec{\theta}}(\cdot|\vec{x})$ and $q_{\vec{\theta}}(\cdot|\vec{x}^{\prime})$ as a function of $\vec{\delta}$ using the logistic model with different values of $\vec{\theta}$: (a) Parameters minimizing the natural misclassification error probability $P_e$, (b) Parameters minimizing the adversarial misclassification error probability $P_e^{\prime}$.}
	\label{fig:frd_logistic}
\end{figure}

Let us now consider the \textsc{FRD} of the two models around the point $\vec{x} = \vec{0}$, which gives a point that lies in the decision boundary, by letting $\vec{\delta} = \vec{x}^{\prime} - \vec{x}$ vary in the $\ell_{\infty}$ ball $\mathcal{B}_{\infty,\varepsilon} = \{ \vec{\delta} : \| \vec{\delta} \|_{\infty} \le \varepsilon \}$, with $\varepsilon = 0.1$. Fig. \ref{fig:frd_nat} displays the \textsc{FRD} for the parameters $\vec{\theta}$ which minimize the misclassification error probability $P_e$, and Fig.~\ref{fig:frd_adv} shows the \textsc{FRD} for the parameters $\vec{\theta}$ which minimize the adversarial misclassification error probability $P_e^{\prime}$. Clearly, the abrupt transition of $q_{\vec{\theta}}(1|\vec{x})$ in Fig.~ \ref{fig:statistical_manifold_nat} corresponds to a sharp increase on the \textsc{FRD} as $\|\vec{\delta} \|_{\infty}$ increases. On the contrary, for a flatter manifold as in Fig.~\ref{fig:statistical_manifold_adv}, the \textsc{FRD} increases much more slowly as $\|\vec{\delta} \|_{\infty}$ increases. This example shows how  \textsc{FRD} reflects the shape of the statistical manifold $\mathcal{C}$. 

Our goal in this work is to use the \textsc{FRD} to control the shape of the statistical manifold by regularizing the misclassification risk.

The rest of this section is organized as follows. We begin by introducing the FIRE risk function, which is our main theoretical proposal to improve the robustness of neural networks.  We continue with the evaluation of the \textsc{FRD} given by expression \eqref{eq:rao_def} for the binary and multiclass classification frameworks and provide some exciting properties and connections with other standard distances and well-known information divergences.

\subsection{The \textsc{Fire} risk function} \label{sec:fire_risk}


The main proposal of this paper is the \textsc{Fire} risk function, defined as follows:
{ 
\begin{align} 
L_{\textrm{\textsc{Fire}}}(\vec{\theta})  \doteq  \, \E_{p(\vec{x},y)}\Big[& \max_{\vec{x}^{\prime} : \|\vec{x}^{\prime} - \vec{x} \|_{p}\leq \varepsilon}-\log q_{\vec{\theta}}(y|\vec{x}) \nonumber \\ & + \lambda \cdot  d_{R,\mathcal{C}}^2(q_{\vec{\theta}}\big(\cdot|\vec{x}), q_{\vec{\theta}}(\cdot|\vec{x}^\prime)\big) \Big], \label{eq:regularized_risk_fire}
\end{align}}
where $\lambda > 0$ controls the trade-off between natural accuracy and robustness to the adversary. 

\begin{figure}[t!]
\centering
\begin{subfigure}[t]{0.5\columnwidth}
	\centering
	\includegraphics[height=1.3in]{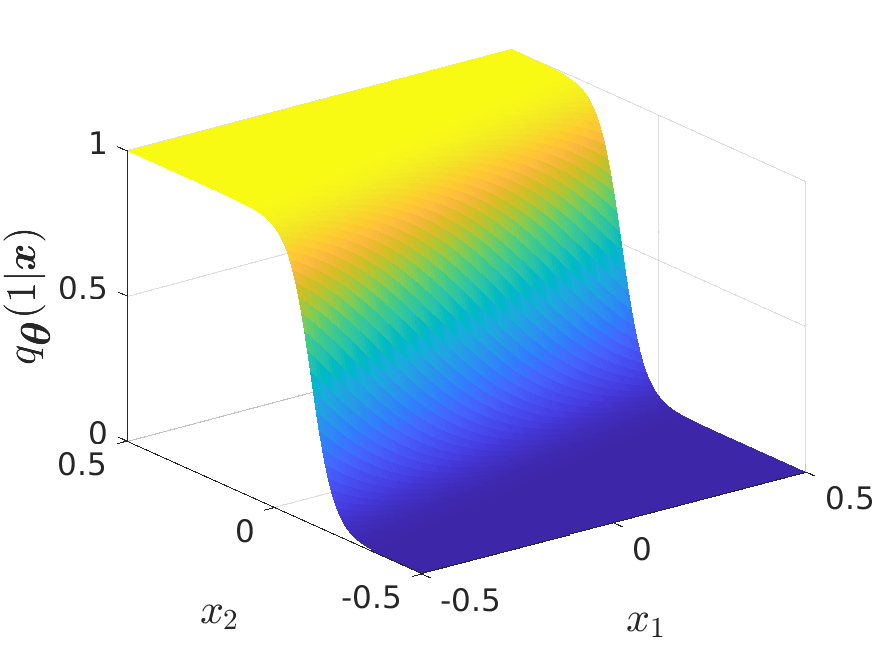}
	\caption{$\lambda = 0$.}
	\label{fig:statistical_manifold_low_lambda}
    \end{subfigure}%
	~ 
	\begin{subfigure}[t]{0.5\columnwidth}
		\centering
		\includegraphics[height=1.3in]{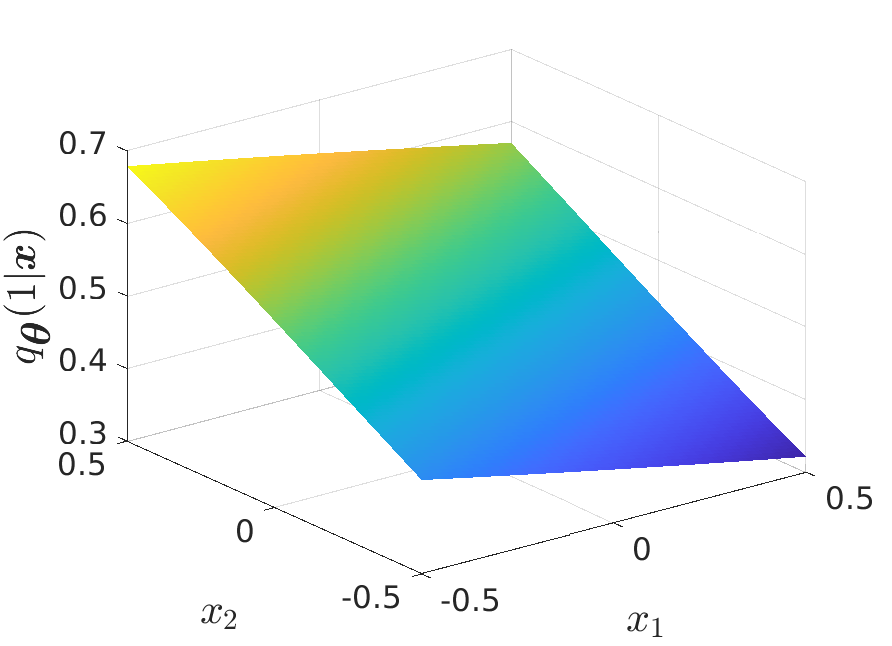}
		\caption{$\lambda = 3.8957$.}
		\label{fig:statistical_manifold_medium_lambda}
	\end{subfigure}
	~
	\begin{subfigure}[t]{0.5\columnwidth}
		\centering
		\includegraphics[height=1.3in]{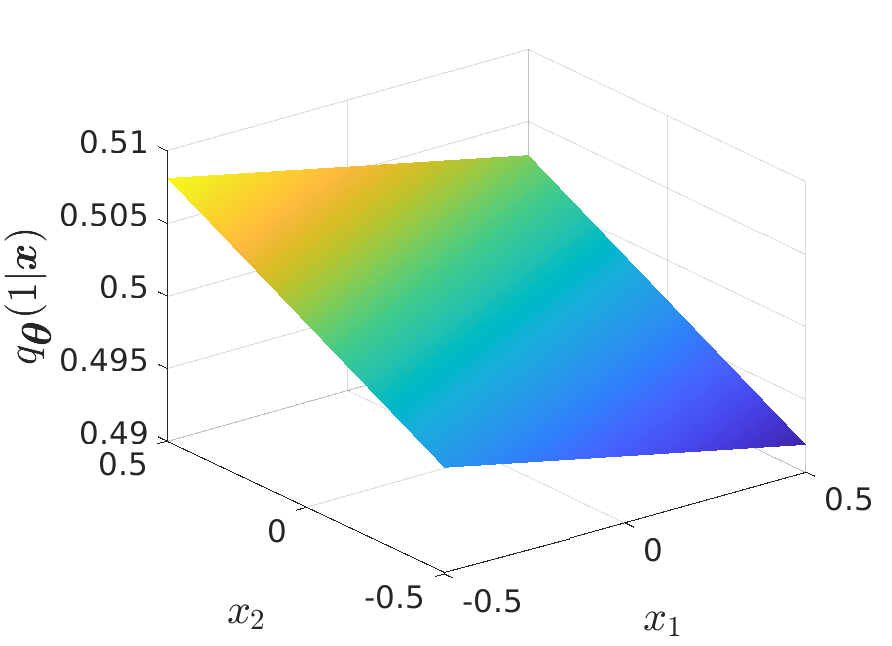}
		\caption{$\lambda = 98.1730$.}
		\label{fig:statistical_manifold_high_lambda}
	\end{subfigure}
	\caption{Visualization of statistical manifold $\mathcal{C}$ defined by the model $q_{\vec{\theta}}(y|\vec{x}) = 1/[1 + \exp(- y \, \vec{\theta}^{\intercal} \vec{x})]$ when minimizing the \textsc{Fire} risk function for different values of $\lambda$: (a) No adversarial \textsc{FRD} regularization, (b) Medium adversarial \textsc{FRD} regularization, (c) High adversarial \textsc{FRD} regularization.}
\label{fig:statistical_manifolds_fire}
\end{figure}

In Fig. \ref{fig:statistical_manifolds_fire}, we show the shape of the statistical manifold $\mathcal{C}$ as $\lambda$ is varied for the logistic regression model discussed in Section \ref{sec:motivation} (see also Section \ref{sec:gaussian_example}). Notice that when no \textsc{FRD} regularization is used, the manifold in Fig. \ref{fig:statistical_manifold_low_lambda} is very similar to the one in Fig. \ref{fig:statistical_manifold_nat}. As the value of $\lambda$ increases, the weight of the \textsc{FRD} regularization term also increases.  As a consequence, the statistical manifold is flattened as expected which is illustrated in Fig. \ref{fig:statistical_manifold_medium_lambda}. However, as shown in Fig. \ref{fig:statistical_manifold_high_lambda}, setting $\lambda$ to a very high value causes the statistical manifold to become extremely flat. This means that the model is basically independent of the input, and the classification performance will be poor. Notice the similarities between Fig. \ref{fig:statistical_manifolds} and Fig. \ref{fig:statistical_manifolds_fire}.

In what follows, we derive closed-form expressions of the \textsc{FRD} for general classification problems. However, for the sake of clarity, we begin with the binary case. 

\subsection{\textsc{FRD} for the case of binary classification} \label{sec:binary_frd}

Let us first consider the binary classification setting, in which $\mathcal{X} \subseteq \R^n$ and  $\mathcal{Y} = \{-1, 1\}$ are input and  label spaces, respectively. Consider an arbitrary given model:  
\begin{equation} \label{eq:binary_model} 
q_{\vec{\theta}}(y|\vec{x}) = \frac{1}{1+e^{-h_{\vec{\theta}}(\vec{x})y}}. 
\end{equation}
Here $h_{\vec{\theta}}(\vec{x})$ represents an arbitrary parametric representation or latent code of the input  $\vec{x}$. As a matter of fact, we only need to assume that $h_{\vec{\theta}}$ is a smooth function. The \textsc{FRD} for this model can be computed in closed-form, as shown in the following result. The proof is relegated to Section~\ref{AppendixB}.

\begin{figure}[t!]
	\centering
	\begin{subfigure}[t]{0.5\columnwidth}
		\centering
		\includegraphics[height=1.3in]{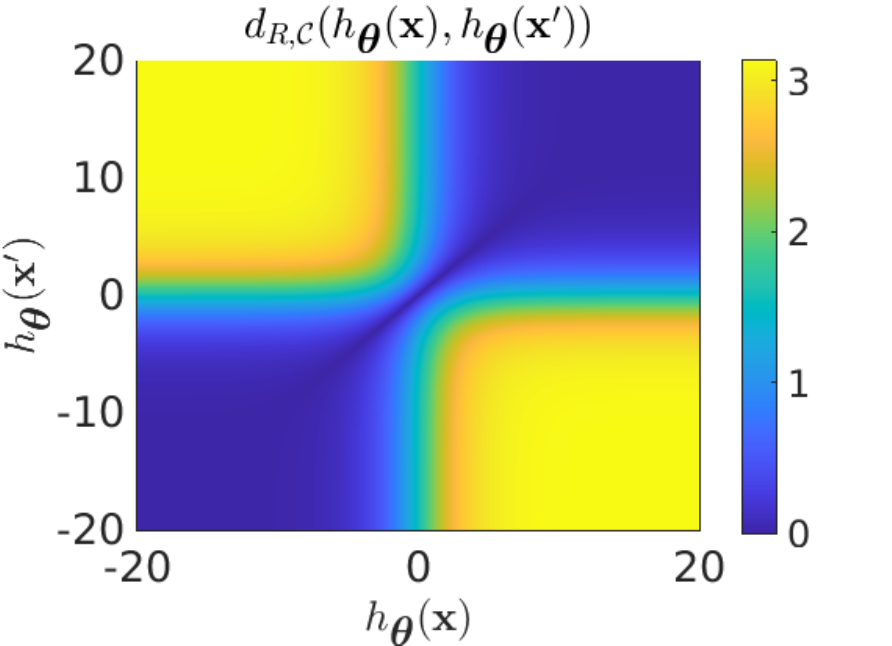}
		\caption{Fisher-Rao distance.}
		\label{fig:rao_binary}
	\end{subfigure}%
	~ 
	\begin{subfigure}[t]{0.5\columnwidth}
		\centering
		\includegraphics[height=1.3in]{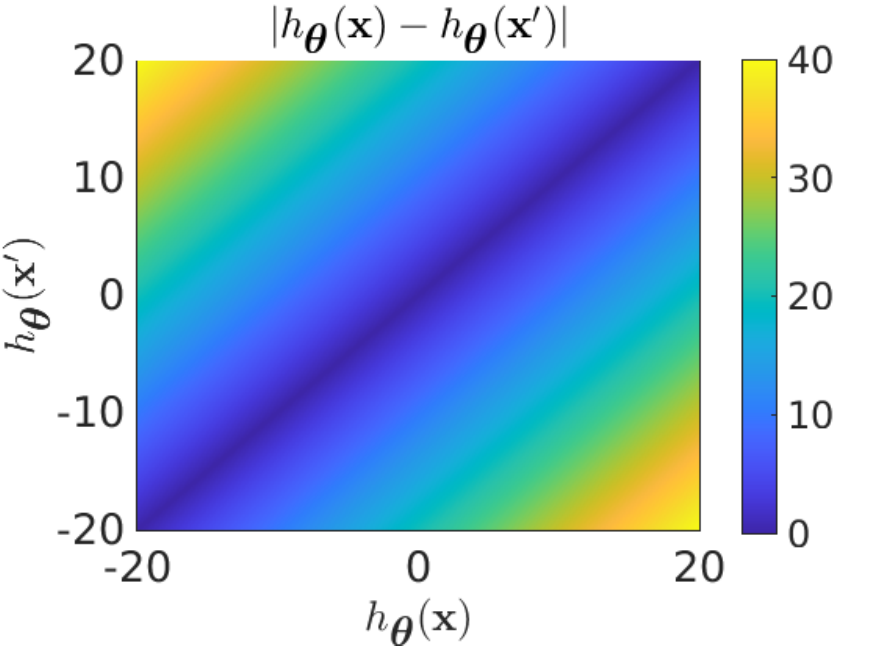}
		\caption{Euclidean distance.}
		\label{fig:euclidean_binary}
	\end{subfigure}
	\caption{Comparison between \textsc{FRD}  and Euclidean distance.}
	\label{fig:binary_distances}
\end{figure}

\begin{theorem}[\textsc{FRD} for binary classifier] \label{thm:binary_frd} The \textsc{FRD} between soft-predictions $q_{\vec{\theta}} \equiv  q_{\vec{\theta}}(\cdot|\vec{x} )$ and $q_{\vec{\theta}}^\prime \equiv  q_{\vec{\theta}}(\cdot|\vec{x}^\prime)$, according to  \eqref{eq:binary_model} and corresponding to inputs $\vec{x}$ and $\vec{x}^\prime$, is given by
\begin{equation} \label{eq:fisher_rao_binary} d_{R,{\mathcal{C}}}(q_{\vec{\theta}}, q_{\vec{\theta}}^\prime) = 2 \Big |\arctan(e^{h_{\vec{\theta}}(\vec{x}^\prime)/2}) 
	- \arctan(e^{h_{\vec{\theta}}(\vec{x})/2})\Big|. 
	\end{equation}
\end{theorem}
For illustration purposes, Fig. \ref{fig:rao_binary} shows the behavior of the \textsc{FRD} with respect to changes in the latent code compared with the Euclidean distance. It can be observed that the resulting  \textsc{FRD} is rather sensitive to variations in the latent space  when $h_{\vec{\theta}}(\vec{x}) \approx 0$ while being close to zero for the region in which $|h_{\vec{\theta}}(\vec{x})|$ is large and $|h_{\vec{\theta}}(\vec{x'})| \ll |h_{\vec{\theta}}(\vec{x})|$. This asymmetric behavior is in sharp contrast with the one of the Euclidean distance. However, these quantities are related as shown by the next proposition.


\begin{proposition}[\textsc{FRD} vs. Euclidean distance] \label{thm:fisher_rao_bound_binary} The Fisher-Rao distance can be bounded as follows:
	\begin{equation} d_{R,{\mathcal{C}}}(q_{\vec{\theta}}, q_{\vec{\theta}}^\prime) \le \frac{1}{2} \big|h_{\vec{\theta}}(\vec{x}^\prime) - h_{\vec{\theta}}(\vec{x})\big|,	\end{equation}
	for any pair of inputs $\vec{x}, \vec{x}^\prime \in\mathcal{X}$. 
\end{proposition}
The proof of this proposition is relegated to Section~\ref{AppendixC}. 

\subsubsection{Logistic regression:} A particular case of significant importance is that of logistic regression: $h_{\vec{\theta}}(\vec{x}) = \vec{\theta}^{\intercal} \vec{x}$. In this case, the \textsc{FRD} reduces to:
\begin{align}
d_{R,\mathcal{C}}(q_{\vec{\theta}},q_{\vec{\theta}}^\prime) = 2 \left| \arctan ( e^{\vec{\theta}^{\intercal} \vec{x}^{\prime}/2}) - \arctan( e^{\vec{\theta}^{\intercal} \vec{x}/2}) \right|. 
\end{align}
A first-order Taylor approximation in the variable $\vec{\delta} = \vec{x}^{\prime} - \vec{x}$ 
and maximization over $\vec{\delta}$ such that $\| \vec{\delta} \| \le \varepsilon$ yields 
\begin{equation} d_{R,\mathcal{C}}(q_{\vec{\theta}},q_{\vec{\theta}}^\prime) \approx \frac{1}{2 \cosh(\vec{\theta}^{\intercal} \vec{x}/2)} \varepsilon \| \vec{\theta} \|_*, \end{equation}
where $\| \cdot \|_{*}$ is the dual norm of $\| \cdot \|$, which is defined as $\| \vec{z} \|_* \doteq \sup\{ |\vec{z}^{\intercal} \vec{w}| : \| \vec{w} \| \le 1 \}$. Therefore, in this case, we obtain a weighted dual norm regularization on $\vec{\theta}$, with the weighting being large when $\vec{\theta}^{\intercal} \vec{x}$ is close to zero (i.e., points with large uncertainty in the class assignment), and being small when $|\vec{\theta}^{\intercal} \vec{x}|$ is large (i.e., points with low uncertainty in the class assignment). An even more direct connection between the \textsc{FRD} and the dual norm regularization on $\vec{\theta}$ can be obtained from Proposition \ref{thm:fisher_rao_bound_binary}, which leads to
\begin{align} \label{eq:link_lp_rao}
d_{R,\mathcal{C}}(q_{\vec{\theta}},q_{\vec{\theta}}^\prime) \le \frac{1}{2} |\vec{\delta}^{\intercal} \vec{\theta}| \le \frac{1}{2} \varepsilon \| \vec{\theta} \|_*.
\end{align}
Eq. (\ref{eq:link_lp_rao}) formalizes our intuitive idea that $\ell_p$-regularized classifiers tend to be more robust. This is in agreement  with other results, e.g., \cite{torkamani14}.

\subsection{\textsc{FRD} for the case of multiclass classification}

Consider the general $M$-classification problem in which $\mathcal{Y} = \{1, \dots, M\}$, and let 
\begin{equation} \label{eq:standard_model} 
q_{\vec{\theta}}(y|\vec{x}) =  \frac{e^{h_y(\vec{x},\vec{\theta})}}{\sum_{y^\prime \in \mathcal{Y}} e^{h_{y^\prime}(\vec{x},\vec{\theta})}}, 
\end{equation}
be a standard softmax output, where $\vec{h}: \mathcal{X} \times \Theta \to \R^M$ is a parametric representation function and $z_y$ denotes the $y$-th component of the vector $\vec{z}$. The \textsc{FRD} for this model can also be obtained in closed-form as summarized below. The proof is given in Section~\ref{app:proof_thm_multiclass_frd}. 

\begin{theorem}[\textsc{FRD} multiclass classifier] \label{thm:multiclass_frd} 
The \textsc{FRD} between soft-predictions $q_{\vec{\theta}} \equiv  q_{\vec{\theta}}(\cdot|\vec{x} )$ and $q_{\vec{\theta}}^\prime \equiv  q_{\vec{\theta}}(\cdot|\vec{x}^\prime)$, according to  \eqref{eq:standard_model} and corresponding to inputs $\vec{x}$ and $\vec{x}^\prime$, is given by
\begin{align}  
d_{R,\mathcal{C}}(q_{\vec{\theta}}&,q_{\vec{\theta}}^\prime) =
&2 \arccos \left( \sum_{y \in \mathcal{Y}} \sqrt{q_{\vec{\theta}}(y|\vec{x} ) q_{\vec{\theta}}(y|\vec{x}^\prime)} \right). 
\label{eq:fisher_rao_multiclass}
\end{align}
\end{theorem}
\noindent\emph{Remark.} Although not obvious, the \textsc{FRD} for the multiclass case (Eq. \eqref{eq:fisher_rao_multiclass}) is indeed consistent with the \textsc{FRD} for the binary case (Eq. \eqref{eq:fisher_rao_binary}), i.e., they are equal for the case $M=2$ (for further details the reader is  referred to Section~\ref{app:proof_thm_multiclass_frd}).

\subsection{Comparison between \textsc{FRD} and KL divergence}

The Fisher-Rao distance \eqref{eq:fisher_rao_multiclass} 
has some interesting connections with other distances and information divergences. We are particularly interested in its relation with the KL divergence, which is the adversarial regularization mechanism used in the TRADES method \cite{zhang2019theoretically}. The next theorem summarizes the mathematical connection between these quantities. The proof of this theorem  is relegated to  Section~\ref{AppendixE}. 

\begin{theorem}[Relation between \textsc{FRD} and KL divergence] The \textsc{FRD} between soft-predictions $q_{\vec{\theta}} = q_{\vec{\theta}}(\cdot|\vec{x} )$ and $q_{\vec{\theta}}^\prime = q_{\vec{\theta}}(\cdot|\vec{x}^\prime)$, given by \eqref{eq:fisher_rao_multiclass} is related to the KL divergence through the inequality:
\begin{equation} \label{eq:rao_bound_kl} 1 - \cos\left(\frac{d_{R,\mathcal{C}}(q_{\vec{\theta}} ,q_{\vec{\theta}}^\prime)}{2} \right) \le \frac{1}{2} \text{KL}(q_{\vec{\theta}} ,q_{\vec{\theta}}^\prime), \end{equation}
which means that the KL divergence is a surrogate of the \textsc{FRD}. In addition, it can also be shown that the KL divergence is a second-order approximation of the \textsc{FRD}, i.e.,
\begin{equation} \label{eq:second_order_rao_kl} \text{KL}(q_{\vec{\theta}} \| q_{\vec{\theta}}^\prime) = \frac{1}{2} d_{R,\mathcal{C}}^2(q_{\vec{\theta}},q_{\vec{\theta}}^\prime) + \mathcal{O}(d_{R,\mathcal{C}}^3(q_{\vec{\theta}},q_{\vec{\theta}}^\prime)), \end{equation}
where $\mathcal{O}(\cdot)$ denotes big-O notation. \label{thm:relation_kl_and_frd}
\end{theorem}

The above result shows that the KL is a weak approximation of the \textsc{FRD} in the sense that it gives an upper bound and a second-order approximation of the geodesic distance. However, in general, we are interested in distances over arbitrarily distinct softmax distributions, so it is clear that the KL divergence and the \textsc{FRD} can behave very differently. In fact, only the latter measures the actual distance on the statistical manifold $\mathcal{C} \doteq \big\{ q_{\vec{\theta}}(\cdot|\vec{x}) : \vec{x} \in \mathcal{X} \big\}$ (for further details, the reader is referred to Section \ref{app:review_frd}). In the next section, we show that this has an important consequence on the set of solutions obtained by minimizing the respective empirical risks while varying $\lambda$.






\begin{figure*}
	\centering
	\begin{subfigure}[b]{0.65\columnwidth}
		\centering
		\includegraphics[width=1\columnwidth]{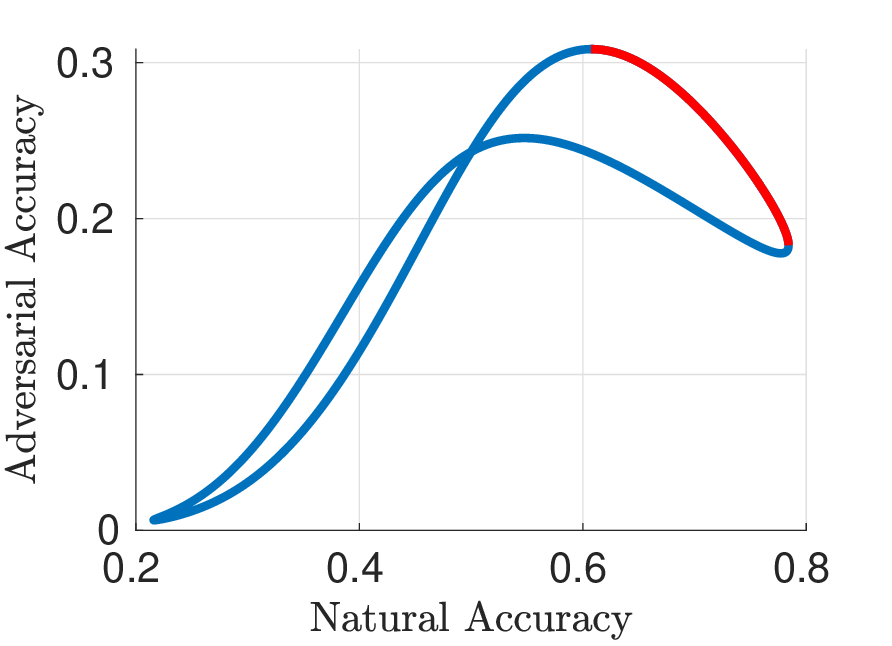}
		\caption{Pareto-optimal points}
		\label{fig:nat_vs_adv_acc_gaussian_pareto}
	\end{subfigure}
	\hfill
	\hspace{-20mm}
	\begin{subfigure}[b]{0.65\columnwidth}
		\centering
		\includegraphics[width=1\columnwidth]{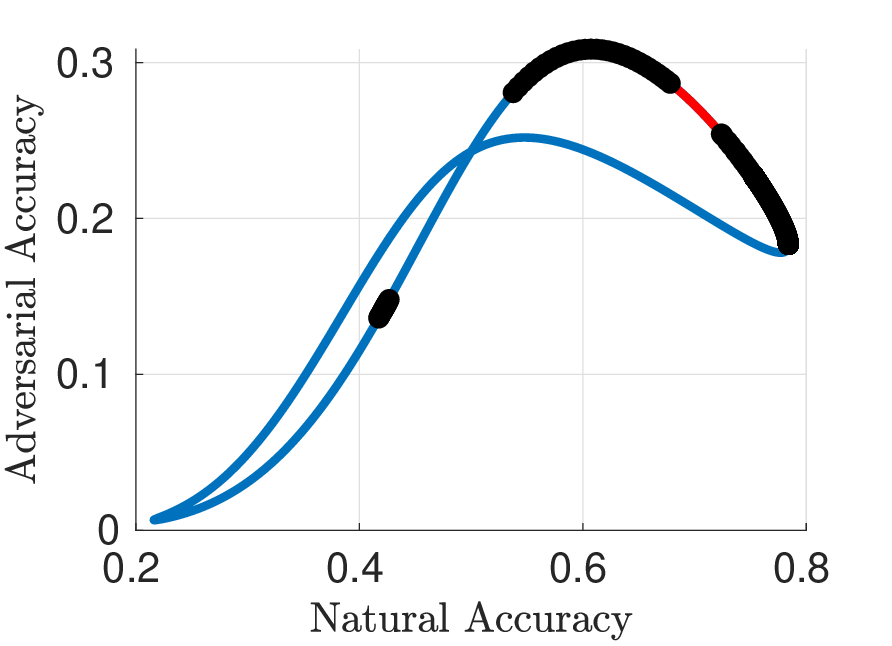}
		\caption{TRADES}
		\label{fig:nat_vs_adv_acc_gaussian_kl}
	\end{subfigure}
	\hfill
		\hspace{-30mm}
	\begin{subfigure}[b]{0.65\columnwidth}
		\centering
		\includegraphics[width=1\columnwidth]{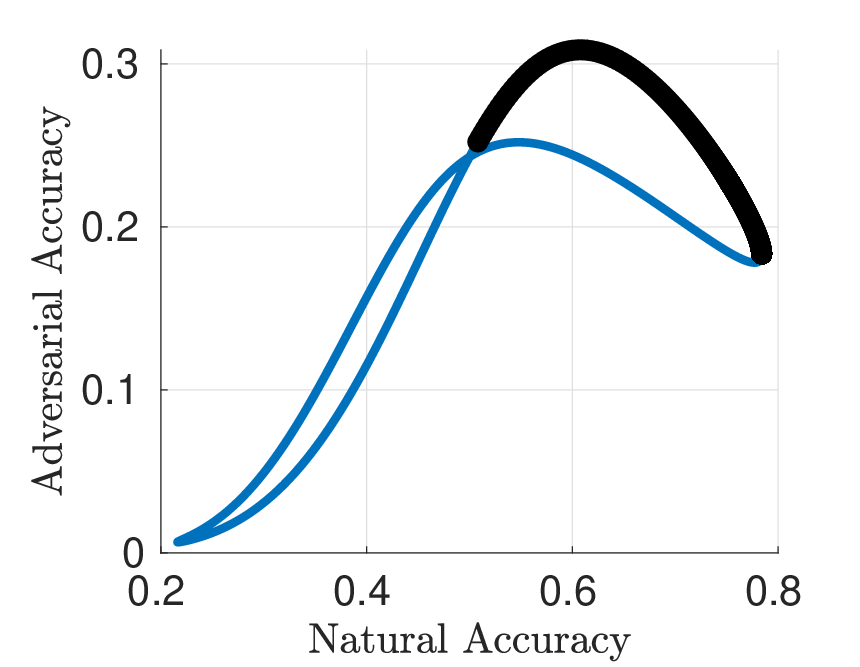}
		\caption{FIRE}
		\label{fig:nat_vs_adv_acc_gaussian_fire}
	\end{subfigure}
	\hfill
	\caption{Plot of all the possible points $(1-P_e(\vec{\theta}), 1-P_e^{\prime}(\vec{\theta}))$ for the Gaussian model with $\varepsilon = 0.1$, $\vec{\mu} = [-0.0218; 0.0425]$ and $\vec{\Sigma} = [ 0.0212, 0.0036; 0.0036, 0.0042]$ shown in blue. In red, we show the Pareto-optimal points (Fig. \ref{fig:nat_vs_adv_acc_gaussian_pareto}). In black, we show the solutions obtained by minimizing the risk  $L_{\textrm{TRADES}}(\vec{\theta})$ in equation \eqref{eq:trades-loss} (Fig. \ref{fig:nat_vs_adv_acc_gaussian_kl}), and the risk  $L_{\textrm{FIRE}}(\vec{\theta})$ in  \eqref{eq:regularized_risk_fire} (Fig. \ref{fig:nat_vs_adv_acc_gaussian_fire}).
	}
	\label{fig:nat_vs_adv_acc_gaussian}
\end{figure*}

\section{Accuracy-Robustness Trade-offs and Learning in the Gaussian Model} \label{sec:gaussian_example}

To illustrate the \textsc{Fire} loss and the role of the Fisher-Rao distance to encourage robustness, we study the natural-adversarial accuracy trade-off for a simple logistic regression and Gaussian model and compare the performance of the predictor trained on the \textsc{Fire} loss with those of ACE and TRADES losses, in equations \eqref{eq:ace_loss} and  \eqref{eq:trades-loss}, respectively. 

\subsection{Accuracy-robustness trade-offs}\label{theory-example}

Consider a binary example with a simplified logistic regression model. Therefore, in this section we assume that $\mathcal{Y} = \{ -1, 1 \}$ and the softmax probability 
\begin{equation} \label{eq:logistic_model} q_{\vec{\theta}}(y|\vec{x}) = \frac{1}{1 + \exp(- y \, \vec{\theta}^{\intercal} \vec{x})}, \end{equation}
We choose the standard adversary obtained by maximizing the cross-entropy loss, i.e.,
\begin{align}  \vec{x}^{\prime*} & = \underset{\vec{x}^\prime : \| \vec{x}^\prime - \vec{x} \| \le \varepsilon}{\text{argmax}} -\log q_{\vec{\theta}}(y|\vec{x}^\prime). \end{align}
For simplicity\footnote{The analysis can be extended to an arbitrary norm but it is somewhat simplified for the 2-norm case.}, in this section we assume that the adversary uses the 2-norm (i.e, $\| \cdot \| = \| \cdot \|_2$). In such case, $\vec{x}^{\prime*}$ can be written as $\vec{x}^{\prime*} = \vec{x} - \varepsilon \, y \, \vec{\theta}/\| \vec{\theta} \|_2.$

We also assume that the classes are equally likely and that the conditional inputs given the class are Gaussian distributions with the particular form $\vec{x}|y \sim \mathcal{N}(y \vec{\mu}, \vec{\Sigma})$. In this case, we can write the natural and adversarial misclassification probabilities as:
\begin{align} \label{eq:pe_nat_gaussian} P_e(\vec{\theta}) & \doteq \P( f_{\vec{\theta}}(\vec{X}) \ne Y) = \Phi \left( \frac{- \vec{\theta}^{\intercal} \vec{\mu}}{\sqrt{\vec{\theta}^{\intercal} \vec{\Sigma} \vec{\theta}}} \right), \\ \label{eq:pe_adv_gaussian} P_e^{\prime}(\vec{\theta}) & \doteq \P( f_{\vec{\theta}}(\vec{X}^{\prime*}) \ne Y) = \Phi \left( \frac{\varepsilon \| \vec{\theta} \|_2 - \vec{\theta}^{\intercal} \vec{\mu}}{\sqrt{\vec{\theta}^{\intercal} \vec{\Sigma} \vec{\theta}}} \right), \end{align}
where $\Phi$ denotes the cumulative distribution function of the standard normal random variable. The following result provides lower and upper bounds for $P_e^{\prime}(\vec{\theta})$ in terms of $\varepsilon$ and the eigenvalues of $\vec{\Sigma}$. Notice that the bounds get sharper as $\varepsilon$ or the spread of $\vec{\Sigma}$ decreases and are tight if $\vec{\Sigma} = \sigma^2 \vec{I}$.

\begin{proposition}[Accuracy-robustness trade-offs] \label{prop-pareto}
	The adversarial misclassification probability $P_e^{\prime}(\vec{\theta})$ satisfies the inequalities:
	\begin{align} \Phi\left( \frac{\varepsilon}{\lambda^{1/2}_{\max}(\vec{\Sigma})} + \Phi^{-1} (P_e(\vec{\theta})) \right) & \le P_e^{\prime}(\vec{\theta}), \\ \Phi\left(  \frac{\varepsilon}{\lambda^{1/2}_{\min}(\vec{\Sigma})} + \Phi^{-1} (P_e(\vec{\theta})) \right) & \ge P_e^{\prime}(\vec{\theta}).\end{align}
\end{proposition}
The proof of this proposition is relegated to Section~\ref{AppendixF}.

\subsection{Learning}

Let us consider the 2-dimensional case, i.e., $n =2$. From expressions \eqref{eq:pe_nat_gaussian} and \eqref{eq:pe_adv_gaussian}, it can be noticed that both $P_e(\vec{\theta})$ and $P_e^{\prime}(\vec{\theta})$ are independent of the 2-norm of $\vec{\theta}$. Therefore, we can parameterize $\vec{\theta}$ as $\vec{\theta} = [\cos(\alpha), \sin(\alpha)]^{\intercal}$ with $\alpha \in [0,2\pi)$ without any loss of generality. Thus, $(1-P_e(\vec{\theta}), 1-P_e^{\prime}(\vec{\theta}))$ for all values of $\alpha$ gives a curve that represents all the possible values of the natural and adversarial accuracies for this setting. Fig. \ref{fig:nat_vs_adv_acc_gaussian_pareto} shows this curve for a particular choice\footnote{In this experiment, we obtained the components in $\vec{\mu}$ by sampling  $\mathcal{N}(0,1/400)$. We also defined a matrix $\vec{A}$ with samples of the same distribution and constructed $\vec{\Sigma}$ as $\vec{\Sigma} = \vec{A} \vec{A}^{\intercal}$.} of $\varepsilon$, $\vec{\mu}$, and $\vec{\Sigma}$. As can be observed, the solution which maximizes the natural accuracy gives poor adversarial accuracy and viceversa\footnote{The (normalized) value of $\vec{\theta}$ that maximizes $1-P_e(\vec{\theta})$ is $\theta_{\text{nat}}^* = \vec{\Sigma}^{-1} \vec{\mu} / \| \vec{\Sigma}^{-1} \vec{\mu} \|_2$ (see, for instance, \cite{bishop2006}) and corresponds to $\alpha_{\text{nat}}^* \approx 1.814$, giving $1-P_e(\vec{\theta}) \approx 0.784$ and $1-P_e^{\prime}(\vec{\theta}) \approx 0.183$. The value of $\theta$ that maximizes $1-P_e^{\prime}(\vec{\theta})$ is $\alpha_{\text{adv}}^* \approx 2.783$, giving $1-P_e(\vec{\theta}) \approx 0.608$ and $1-P_e^{\prime}(\vec{\theta}) \approx 0.309$.}. The set of Pareto-optimal points (i.e., the set of points for which there is no possible improvement in terms of both natural and adversarial accuracy or, equivalently, the set $\{\max_{\vec{\theta}} \; \beta(1 - P_e(\vec{\theta})) + (1-\beta)(1- P_e'(\vec{\theta})) : 0 \le \beta \le 1 \}$) are also shown in Fig. \ref{fig:nat_vs_adv_acc_gaussian_pareto}. In particular, this set contains the Maximum Average Accuracy (MAA) given by
\begin{equation} \label{eq:maa} \text{MAA} \doteq \underset{\vec{\theta}\in\Theta }{\text{max}} \; \left( 1 - \frac{P_e(\vec{\theta}) + P_e^{\prime}(\vec{\theta})}{2} \right). \end{equation}
This is a metric of particular importance, which combines with equal weights both  natural and adversarial accuracies.

In addition, we present the solution of the (local) Empirical Risk Minimization (ERM)\footnote{For experiments, we used $10^4$ samples for each different class and a BFGS Quasi-Newton method for optimization. The initial value of $\vec{\theta}$ is zero for both TRADES and FIRE. We do not report the result using ACE risk in \eqref{eq:ace_loss} because in this setting we would obtain the trivial solution $\vec{\theta} = \vec{0}$, which is a minimizer of $L_{\textrm{ACE}}(\vec{\theta})$.} for the TRADES risk function as defined in \eqref{eq:trades-loss} for different values of $\lambda$ in Fig. \ref{fig:nat_vs_adv_acc_gaussian_kl}. As can be seen, the curve obtained in the $(1-P_e(\vec{\theta}), 1-P_e^{\prime}(\vec{\theta}))$ space covers a large part of the Pareto-optimal points expect for a segment for which the solution does not behave well. Finally, in Fig. \ref{fig:nat_vs_adv_acc_gaussian_fire} we present the result for the proposed \textsc{Fire} risk function in \eqref{eq:regularized_risk_fire} for different values of $\lambda$. In this case, we observed that the curve of solutions in the $(1-P_e(\vec{\theta}), 1-P_e^{\prime}(\vec{\theta}))$ space covers all the Pareto-optimal points. Moreover, we have observed that, for some $\lambda$, the $\vec{\theta}$ which minimizes the \textsc{Fire} risk matches the $\vec{\theta}$ which achieves the MAA defined in \eqref{eq:maa}, while TRADES method fails in achieving this particularly relevant point. It should be added that none of the methods cover exactly the set of Pareto-optimal points, which is expected, since all loss functions can be considered surrogates for the quantity $\beta P_e(\vec{\theta}) + (1-\beta) P_e'(\vec{\theta})$, where $0 \le \beta \le 1$.
{  We performed a similar comparison between FRD and KL using standard datasets. The results are reported in Appendix \ref{app:rao_vs_kl}}

\begin{table*}[t]
	\tiny
	\centering
	\caption{Comparison between KL and Fisher-Rao based regularizer  under white-box $l_{\infty}$ threat model. \textcolor{black}{Note that we do not use the same hyperparameters as presented in \cite{zhang2019theoretically} for the TRADES method.}}
	\resizebox{0.84\textwidth}{!}{
		\begin{tabular}{c||c|c|c||c|c||c||c}
			\toprule
			
			Defense & Dataset &$\varepsilon$& Structure & Natural  & AutoAttack & { Avg. Acc.} & RunTime\\
			\midrule
			\midrule
			
			TRADES & & & CNN & \textbf{ 99.27 $\pm$ 0.03} & {  94.27 $\pm$ 0.18} & { 96.77 $\pm$ 0.09} & 2h22 \\
			\rowcolor{Gray} FIRE & \multirow{-2}{*}{MNIST}&\multirow{-2}{*}{$0.3$} & CNN & {  99.22 $\pm$ 0.02} & \textbf{  94.44 $\pm$ 0.14} & \textbf{ 96.83 $\pm$ 0.10} & \textbf{2h06} \\
			\midrule
			
			TRADES &  & & WRN-34-10 & { 85.84 $\pm$ 0.31} &  { 50.47 $\pm$ 0.36} & { 68.15 $\pm$ 0.23} & 13h49 \\
			\rowcolor{Gray} FIRE & \multirow{-2}{*}{CIFAR-10} & \multirow{-2}{*}{$8/255$}& WRN-34-10 & \textbf{{ 85.98 $\pm$ 0.09}}  & \textbf{{ 51.45 $\pm$ 0.32}} & \textbf{{ 68.72 $\pm$ 0.22}} & \textbf{11h00}\\
			\midrule
			
			TRADES & & & WRN-34-10 & {  59.62 $\pm$ 0.42} & {  25.89 $\pm$ 0.26}  & {  42.76 $\pm$ 0.26} & 13h49 \\
			\rowcolor{Gray}  FIRE & \multirow{-2}{*}{CIFAR-100} & \multirow{-2}{*}{$8/255$}& WRN-34-10 & \textbf{{  61.03 $\pm$ 0.21}} &  \textbf{{ 26.42 $\pm$ 0.21}} & \textbf{{  43.73 $\pm$ 0.12}}  & \textbf{11h10} \\
			
			\bottomrule
	\end{tabular}}
	\label{tab:comparison_trades}
\end{table*}

\begin{table*}[t]
	\tiny
	\centering
	\caption{Test robustness on different datasets under white-box $l_{\infty}$ attack. { We ran all methods on 5 different tries and reported the mean and standard deviation. The codes for UAT and Atzmon et al. are not publicly available.} { Note that retraining the SOTA methods modifies slightly the experimental results.}}
	\resizebox{0.84\textwidth}{!}{
		\begin{tabular}{c||c|c|c||c|c||c||c}
			\toprule
			
			Defense & Dataset& $\varepsilon$ & Structure & Natural & AutoAttack & { Avg. Acc.} & Runtime\\
			\midrule
			\midrule
			\multicolumn{8}{c}{\textbf{Without Additional Data } }\\
			\midrule
			
			Madry et al. \cite{madry2018} & \multirow{4}{*}{MNIST} &\multirow{4}{*}{$0.3$} & CNN & { 98.53 $\pm$ 0.06} &  { 88.62 $\pm$ 0.23} & {  93.58 $\pm$ 0.14} &\textbf{2h03} \\
			Atzmon et al.\cite{atzmon2019controlling} & & & CNN & \textbf{99.35} & 90.85 & 95.10 & - \\
			TRADES \cite{zhang2019theoretically} & & & CNN & { 99.27 $\pm$ 0.03} & {  94.27 $\pm$ 0.18} & { 96.77 $\pm$ 0.09} & 2h22 \\
			\rowcolor{Gray} FIRE & &  & CNN & {  99.22 $\pm$ 0.02} & \textbf{  94.44 $\pm$ 0.14} & \textbf{ 96.83 $\pm$ 0.10} & 2h06\\ 
			\midrule
			
			Madry et al. \cite{madry2018}& & & WRN-34-10 & \textbf{ 87.56$\pm$ 0.09}  & { 44.07 $\pm$ 0.27} &{ 65.82 $\pm$ 0.15 } &\textbf{10h51} \\
			
			Self Adaptive\cite{huang2020self} & & & WRN-34-10 & {  83.39 $\pm$ 0.19} & {  53.11 $\pm$ 0.29} & {  68.25 $\pm$ 0.14} & 13h57 \\
			
			TRADES \cite{zhang2019theoretically} & & & WRN-34-10 & {  \textcolor{black}{84.79 $\pm$ 0.24}}  & \textcolor{black}{ 52.12$\pm$ 0.28} & \textcolor{black}{ 68.45 $\pm$ 0.12} & 17h49 \\
			
			\rowcolor{Gray}{ FIRE + Self Adaptive}  & & & { WRN-34-10} & {  83.70 $\pm$ 0.36} & {  \textbf{53.26 $\pm$ 0.19}} & {  68.48 $\pm$ 0.13} & {  11h12} \\
			
			Overfitting \cite{rice2020overfitting} & & & WRN-34-10 & {  85.64 $\pm$ 0.55} & {  51.72 $\pm$ 0.56} & {  68.68 $\pm$ 0.44} & 42h01 \\
			\rowcolor{Gray} FIRE & \multirow{-6}{*}{CIFAR-10} & \multirow{-6}{*}{$8/255$}& WRN-34-10 & { 85.98 $\pm$ 0.09}  & { 51.45 $\pm$ 0.32} & \textbf{{ 68.72 $\pm$ 0.22}} & 11h00\\
			
			\midrule
			
			Overfitting \cite{rice2020overfitting} & & & RN-18 & 53.83  & 18.95 & 36.39 & - \\
			Overfitting \cite{rice2020overfitting} &  &&  WRN-34-10 & { 59.22 $\pm$ 0.61} & { 25.99 $\pm$ 0.51} & { 42.61 $\pm$ 0.28} & 42h08 \\
			\rowcolor{Gray}  FIRE &\multirow{-3}{*}{CIFAR-100} &\multirow{-3}{*}{$8/255$} & WRN-34-10 & \textbf{{ 61.03 $\pm$ 0.21}} & \textbf{{ 26.42 $\pm$ 0.21}} & \textbf{{ 43.73 $\pm$ 0.12}} & \textbf{11h10}\\

			

			\midrule 
			\midrule
			\multicolumn{8}{c}{\textbf{With Additional Data Using  80M-TI} }\\
			\midrule
			Pre-training\cite{hendrycks2019using}& \multirow{5}{*}{CIFAR-10} &\multirow{5}{*}{$8/255$} & WRN-28-10 & {  86.93 $\pm$ 0.79}  & {  53.35 $\pm$ 0.81} & {  70.14 $\pm$ 0.54} &  {40h00 + 0h20} \\  
			UAT \cite{alayrac2019labels} & & & WRN-106-8 & 86.46   & 56.03 & 71.24 & - \\
			
			MART \cite{wang2019improving} & & & WRN-28-10 & {  87.39 $\pm$ 0.12}  & {  56.69 $\pm$ 0.28} & {  72.04 $\pm$  0.15} & \textbf{ 13h53}  \\
			RST-adv \cite{carmon2019unlabeled} & & & WRN-28-10 & {  89.49 $\pm$ 0.41} &  {  59.69 $\pm$ 0.26} & {  74.59 $\pm$ 0.27} & 22h12  \\
			\rowcolor{Gray}FIRE & & & WRN-28-10 & \textbf{{  89.73 $\pm$ 0.04}} & \textbf{{  59.97 $\pm$ 0.11}} & \textbf{{  74.86 $\pm$ 0.05}} & 18h30 \\
			
			\bottomrule
	\end{tabular}}
	\label{tab:test_results}
\end{table*}

\section{Experimental Results} \label{sec:experiments}
In this section, we assess our proposed \textsc{Fire} loss' effectiveness to improve neural networks' robustness.\footnote{Codes are available on GitHub at \url{https://github.com/MarinePICOT/Adversarial-Robustness-via-Fisher-Rao-Regularization}}

\subsection{Setup}

\textbf{Datasets : } We resort to standard benchmarks. First, we use MNIST \cite{lecun2010mnist}, composed of 60,000 black and white images of size 28$\times$28 - 50,000 for training, and 10,000 for testing - divided into 10 different classes. Then, we test on CIFAR-10, and CIFAR-100 \cite{Krizhevsky09learningmultiple}, composed of 60,000 color images of size 32$\times$32$\times$3 - 50,000 for training and 10,000 for testing - divided into 10 and 100 classes, respectively. Finally, we also test on CIFAR-10 with additional data thanks to 80 Million Tiny Images \cite{torralba2008tinyimages}, an experiment that will be detailed later.

\textbf{Architectures : } In order to provide fair comparisons, we use standard model architectures. For the MNIST simulations, we use the 7-layer CNN as in \cite{zhang2019theoretically}. For CIFAR-10 and CIFAR-100, we use a WideResNet (WRN) with 34 layers, and a widen factor of 10 (shortened as WRN-34-10) as in \cite{zhang2019theoretically,madry2018, wu2020adversarial}. For the simulations with additional data, we use a WRN-28-10 as in \cite{carmon2019unlabeled}.

\textbf{Training procedure:}   For all standard experiments (without additional data), we use a Stochastic Gradient Descent (SGD) optimizer with a momentum of $0.9$, a weight decay of $5 \cdot 10^{-4}$ { and Nesterov momentum}. We train our models on $100$ epochs with a batch size equal to $256$. The initial learning rate is set to $0.01$ for MNIST and $0.1$ for CIFAR-10 and CIFAR-100. Following \cite{zhang2019theoretically}, the learning rate is divided by $10$ at epochs $75$ and $90$. For our experiments with additional data,  we follow the protocol introduced by \cite{carmon2019unlabeled}, {  using} a cosine decay \cite{loshchilov2016sgdr}, and { training} on $200$ epochs. 

\textbf{Generation of adversarial samples:} 
For all experiments, we use PGD \cite{madry2018} to generate the adversarial examples during training. The loss which is maximized during the PGD algorithm is the Fisher-Rao distance (\textsc{FRD}) {  for our experiments, and the Kullback-Leibler divergence for the TRADES method}. For MNIST, we use 40 steps and 10 steps for the rest. The step size is set to 0.01 for MNIST and 0.007 for the others. The maximal distortion in $l_{\infty}$-norm $\varepsilon$ allowed is $0.031$ for CIFAR-10 and CIFAR-100, and $0.3$ for MNIST. This setting is used in most methods, such as \cite{goodfellow2014,carmon2019unlabeled}.

\textbf{Additional data:} To improve performance, \cite{carmon2019unlabeled} propose the use  of additional data when training on CIFAR-10. Specifically, \cite{carmon2019unlabeled} use 500k additional images from 80M-TI \footnote{Images available at \url{https://github.com/yaircarmon/semisup-adv}}. Those images have been selected such that their $l_{2}$-distance to images from CIFAR-10 are below a threshold. 

\textbf{Hyperparameters:} 
The Rao regularizer used to improve the robustness of neural networks introduces a hyperparameter $\lambda$ to balance natural accuracy and adversarial robustness. We select $\lambda$ { = 12} from the CIFAR-10 simulations and use this value for all the other datasets. Further study on the effect of $\lambda$ is provided in Section \ref{sec:ablations}.

\textbf{Test metrics:} 
First, we provide the accuracy of the model on clean samples after adversarial training (Natural). Second, we test our models using the recently introduced AutoAttack \cite{croce2020reliable}, keeping the same hyperparameters as the ones used in the original paper and code\footnote{The AutoAttack code is available on \url{https://github.com/fra31/auto-attack}}. AutoAttack tests the model under a comprehensive series of attacks and provides a more reliable assessment of robustness than the traditionally used PGD-based evaluation. Given that we care equally about natural and adversarial accuracies, we also compute the average sum of the two, i.e., the Average Accuracy ({ Avg. Acc.}). This is an empirical version of the Maximum Average Accuracy (MAA) defined in  \eqref{eq:maa}. Finally, we report the runtime of each method as the time required to complete the adversarial training. To provide fair comparisons between runtimes, we run the official code of each method on the same $4$ NvidiaV100 GPUs (for further details, see Section \ref{sec:comparison_sota}).



\subsection{Experimental results}

\subsubsection{Kullback-Leibler versus Fisher-Rao regularizer:} \label{sec:comparison_kl}
To disentangle the influence of different regularizers, we compare the Fisher-Rao-based regularizer to the Kullback-Leibler-based regularizer used in TRADES with the exact same model and hyperparameters (as detailed previously) on CIFAR-10, CIFAR-100, and MNIST datasets. {We used $\lambda = 6$ to train the TRADES method, since it is the value presented in the original paper \cite{zhang2019theoretically}.} The results {  are averaged over 5 tries and} summarized in Table \ref{tab:comparison_trades}. Those results confirm the superiority of the proposed regularizer. {  Specifically, the natural and adversarial accuracies increase up to 1\% each under AutoAttack, improving the trade-off up to $1\%$}.

\subsubsection{Comparison with state-of-the-art}\label{sec:comparison_sota}

We compare \textsc{Fire} with the state-of-the-art methods using adversarial training under $l_{\infty}$-norm attacks. { Due to its effectiveness on adversarial performances, we use the self-adaptive scheme from \cite{huang2020self} along with the FIRE loss to smooth the one-hot labels on CIFAR-10, and report the results under FIRE + Self Adaptive.} 
We do not include the Adversarial Weight Perturbation (AWP) \cite{wu2020adversarial} method since it leverages two networks to increase the robustness of the model. { We trained all methods from the state-of-the art on 5 tries and report the mean the standard variance of the results }in Table \ref{tab:test_results}. {  Note that the codes for UAT and Atzmon et al. are not publicly available, therefore we reported the results available on RobustBench \cite{croce2020robustbench}.}


\textbf{Average Accuracy ({ Avg. Acc.}):}
Overall, \textsc{Fire} exhibits the best { Avg. Acc.} among compared methods in all settings. On CIFAR-10, at equivalent method, our \textsc{Fire} method outperforms \cite{huang2020self}. { The gain on natural examples is close to 0.3\% and 0.15\% in adversarial performances, giving an improvemnt of 0.23\% of Avg. Acc.} Moreover, our \textsc{Fire} method performs slightly better (0.04\%) than \cite{rice2020overfitting} but, given that their method requires four times \textsc{Fire}'s runtime to complete its training, the gain of our method appears to be more significant. Besides, \textsc{Fire} outperforms \cite{rice2020overfitting} on the more challenging CIFAR-100 in both { Avg. Acc.}, with more significant gain ({ 1.12\%}) and runtime. Taking all metrics into account, \textsc{Fire} appears to be the best overall method.


\textbf{Runtimes: } 
Interestingly, our method exhibits a significant advantage over previous state-of-the-art methods using similar backbones. 
Our method outruns methods with similar performances by 20\% on average. 
We presume that the difference between the different runtimes comes from the fact that our proposed \textsc{Fire} loss comprises  5 different operations. In contrast, the KL divergence, proposed in \cite{zhang2019theoretically}, is composed of 6 operations.

\subsubsection{Ablation studies}\label{sec:ablations}
\begin{figure}[t]
	\begin{center}
		\includegraphics[width=0.45\textwidth]{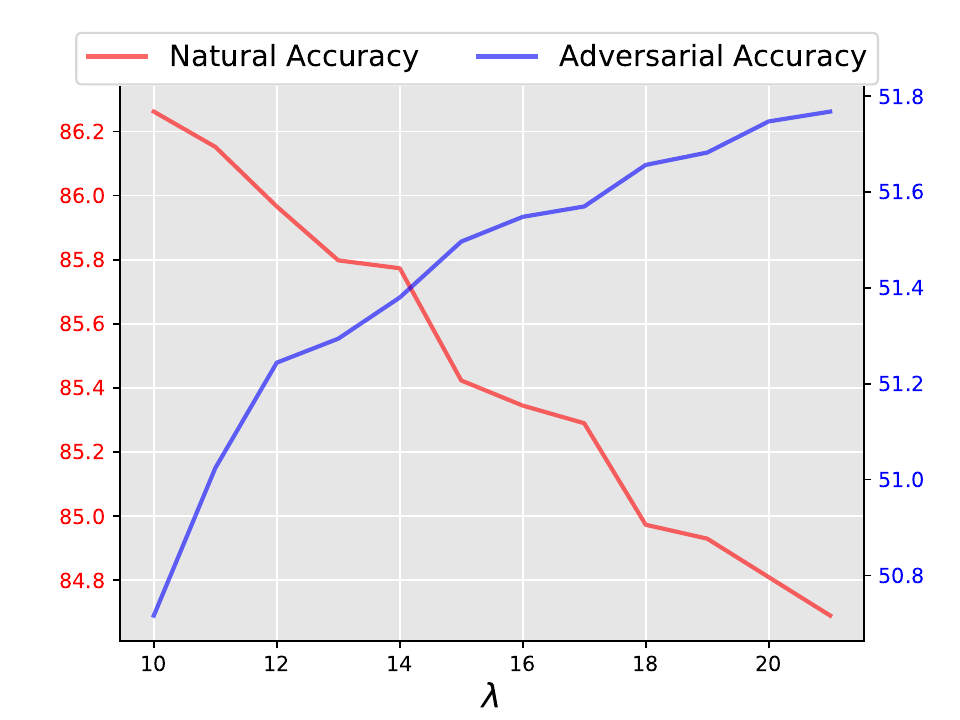}
	\end{center}
	\caption{{ Influence of the hyperparameter $\lambda$ on the natural and adversarial accuracies for \textsc{Fire} regularizer on CIFAR-10.}}
	\label{fig:Influence_lambda}
\end{figure}

\textbf{Influence of $\lambda$:} 
We study the influence of the hyperparameter $\lambda$ on the performances of the \textsc{Fire} method. Figure \ref{fig:Influence_lambda} clearly shows the trade-off between natural and adversarial accuracy under AutoAttack \cite{croce2020reliable}. When increasing $\lambda$, we emphasize our robust regularizer, consequently decreasing the performance on clean samples. Such phenomenon properly aligns with intuition and is also observed in \cite{zhang2019theoretically}. { Even though several values of $\lambda$ lead to reasonable performances, we chose $\lambda=12$ to have a natural accuracy close to the natural accuracy under the TRADES method en CIFAR-10. This ensures a fair comparison. }

\vspace{-2mm}

\section{Proofs of Theorems and Propositions}
\subsection{Review of Fisher-Rao Distance (\textsc{FRD})} \label{app:review_frd}
Consider the family of probability distributions over the classes\footnote{Since we are interested in the dependence of $q_{\vec{\theta}}(\cdot|\vec{x})$ with changes in input $\vec{x}$ and, particularly, its robustness to adversarial perturbations, we consider $\vec{x}$ as the ``parameters'' of the model over which the regularity conditions must be imposed.} \mbox{$\mathcal{C} \doteq \big\{ q_{\vec{\theta}}(\cdot|\vec{x}) : \vec{x} \in \mathcal{X} \big\}$}. Assume that the following regularity conditions hold \cite{atkinson1981}:
\begin{enumerate}[label=(\roman*)]
	\item $\nabla_{\vec{x}} \, q_{\vec{\theta}}(y|\vec{x})$ exists for all $\vec{x},y$ and $\vec{\theta}\in\Theta $;
	\item $\sum_{y \in \mathcal{Y}} \nabla_{\vec{x}} \, q_{\vec{\theta}}(y|\vec{x}) = 0$ for all $\vec{x}$ and $\vec{\theta}\in\Theta $;
	\item $\vec{G}(\vec{x}) = \E_{Y \sim q_{\vec{\theta}}(\cdot|\vec{x})} \big[ \nabla_{\vec{x}} \, \log q_{\vec{\theta}}(Y|\vec{x}) \nabla_{\vec{x}}^{\intercal} \, \log q_{\vec{\theta}}(Y|\vec{x}) \big]$ is positive definite for any $\vec{x}$ and $\vec{\theta}\in\Theta $.
\end{enumerate}
Notice that if (i) holds, (ii) also holds immediately for discrete distributions over finite spaces (assuming that $\sum_{y \in \mathcal{Y}}$ and $\nabla_{\vec{x}}$ are interchangeable operations) as in our case. When (i)-(iii) are met, the variance of the differential form $\nabla_{\vec{x}}^{\intercal} \log q_{\vec{\theta}}(Y|\vec{x}) d\vec{x}$ can be interpreted as the square of a differential arc length $ds^2$ in the space $\mathcal{C}$, which yields 
\begin{equation} \label{eq:ds2} 
d s^2 = \langle d\vec{x}, d\vec{x} \rangle_{\vec{G}(\vec{x})} = d\vec{x}^{\intercal} \vec{G}(\vec{x}) d\vec{x}. 
\end{equation}
Thus, $\vec{G}$, which is the Fisher Information Matrix (FIM), can be adopted as a metric tensor. We now consider a curve $\vec{\gamma}: [0,1] \to \mathcal{X}$ in the input space connecting two arbitrary points $\vec{x}$ and $\vec{x}^\prime$, i.e., such that $\vec{\gamma}(0) = \vec{x} $ and $\vec{\gamma}(1) = \vec{x}^\prime$. Notice that this curve induces the following curve in the space $\mathcal{C}$: $q_{\vec{\theta}}(\cdot|\vec{\gamma}(t))$ for $t \in [0,1]$. The Fisher-Rao distance between the distributions $q_{\vec{\theta}} = q_{\vec{\theta}}(\cdot|\vec{x})$ and $q_{\vec{\theta}}^\prime = q_{\vec{\theta}}(\cdot|\vec{x}^\prime)$ will be denoted as $d_{R,\mathcal{C}}(q_{\vec{\theta}},q_{\vec{\theta}}^\prime)$ and is formally defined by
\begin{equation} \label{eq:rao_def} 
d_{R,\mathcal{C}}(q_{\vec{\theta}},q_{\vec{\theta}}^\prime) \doteq \underset{\vec{\gamma}}{\text{inf}} \int_{0}^{1} \sqrt{\frac{d \vec{\gamma}^{\intercal}(t)}{dt} \vec{G}(\vec{\gamma}(t)) \frac{d \vec{\gamma}(t)}{dt}}, 
\end{equation}
where the infimum is taken over all piecewise smooth curves. This means that the \textsc{FRD} is the length of the \emph{geodesic} between points $\vec{x}$ and $\vec{x}^\prime$ using the FIM as the metric tensor. In general, the minimization of the functional in \eqref{eq:rao_def} is a problem that can be solved using the well-known Euler-Lagrange differential equations. Several examples for simple families of distributions can be found in \cite{atkinson1981}. 

\subsection{Proof of Theorem \ref{thm:binary_frd}}
\label{AppendixB}

To compute the \textsc{FRD}, we first need to compute the FIM of the family $\mathcal{C}$ with $q_{\vec{\theta}}(y|\vec{x})$ given in expression  \eqref{eq:binary_model}. A direct calculation gives:
\begin{align} 
\vec{G}(\vec{x}) & = \frac{e^{h_{\vec{\theta}}(\vec{x})}}{\left(1 + e^{h_{\vec{\theta}}(\vec{x})} \right)^2} \nabla_{\vec{x}} h_{\vec{\theta}}(\vec{x}) \nabla_{\vec{x}}^{\intercal} h_{\vec{\theta}}(\vec{x}). 
\end{align}	
It is clear from this expression that the FIM of this model is of rank one and therefore singular. This matches the fact that $q_{\vec{\theta}}(y|\vec{x})$ has a single degree of freedom given by $h_{\vec{\theta}}(\vec{x})$. Therefore, the statistical manifold $\mathcal{C}$ has dimension $1$.
	
To proceed, we consider the following model:
\begin{equation} q_{\vec{\theta}}(y|u) = \frac{1}{1+e^{-u y}}, \end{equation}
where we have defined $u = h_{\vec{\theta}}(\vec{x})$. This effectively removes the model ambiguities because $q(y|u) \ne q(y|u^\prime)$ if and only if $ u \ne u^\prime$. Note that $ q_{\vec{\theta}}(y|u)$, for fixed $\vec{\theta}$, can be interpreted as a one-dimensional parametric model with parameter $u$. Its FIM is a scalar that can be readily obtained, yielding:
\begin{equation} G(u) = \frac{e^u}{(1+e^u)^2}. \end{equation}
Clearly, the FIM $G(u)$ is non-singular for any $u$ (i.e., $G(u) > 0$ for any $u$) as required. Let $\mathcal{D} = \{ q_{\vec{\theta}}(\cdot | u) : u \in \mathcal{U} \}$, where $\mathcal{U} = h_{\vec{\theta}}(\mathcal{X})$ and consider two distributions in $\mathcal{D}$: $q_{\vec{\theta}} = q_{\vec{\theta}}(\cdot|u)$ and $q_{\vec{\theta}}' = q_{\vec{\theta}}(\cdot|u^\prime)$. Then, the \textsc{FRD} can be evaluated directly as follows \cite{atkinson1981}[Eq. (3.13)]:
\begin{align} \label{eq:frd_scalar_model} d_{R,\mathcal{D}}(q_{\vec{\theta}},q_{\vec{\theta}}') & = \left| \int_{u}^{u^\prime} G^{1/2}(v) d v \right| \nonumber \\ & = 2 \Big|\arctan\left(e^{u^\prime/2}\right) - \arctan\left(e^{u/2}\right)\Big|. \end{align}
Therefore, the \textsc{FRD} between the distributions $q_{\vec{\theta}} = q_{\vec{\theta}}(\cdot|\vec{x} )$ and $q_{\vec{\theta}}^\prime = q_{\vec{\theta}}(\cdot|\vec{x}^\prime)$ can be directly obtained by substituting $u = h_{\vec{\theta}}(\vec{x})$ and $u^\prime = h_{\vec{\theta}}(\vec{x}')$ in Equation \eqref{eq:frd_scalar_model}, yielding the final result in expression \eqref{eq:fisher_rao_binary}.


\subsection{Proof of Theorem \ref{thm:multiclass_frd}} \label{app:proof_thm_multiclass_frd}


As in the binary case developed in Section \ref{sec:binary_frd}, the FIM of the family $\mathcal{C}$ with $q_{\vec{\theta}}(y|\vec{x})$ given in expression \eqref{eq:standard_model} is singular. To shows this, we first notice that $q_{\vec{\theta}}(y|\vec{x})$ can be written as
\begin{equation} q_{\vec{\theta}}(y|\vec{x}) = \frac{e^{g_y(\vec{x},\vec{\theta})}}{\sum_{y^\prime \in \mathcal{Y}} e^{g_{y^\prime}(\vec{x},\vec{\theta})}}, \end{equation}
where $g_y(\vec{x},\vec{\theta}) = h_y(\vec{x},\vec{\theta}) - h_1(\vec{x},\vec{\theta})$. Since $g_1(\vec{x},\vec{\theta}) = 0$, this shows that  $\mathcal{C}$ has $M-1$ degrees of freedom: $g_2(\vec{x},\vec{\theta}), \ldots, g_M(\vec{x},\vec{\theta})$. A direct calculation of the FIM gives
\begin{align} \vec{G}(\vec{x}) & = \sum_{y=2}^{M} q_{\vec{\theta}}(y|\vec{x}) \nabla_{\vec{x}} g_y(\vec{x},\vec{\theta}) \nabla_{\vec{x}}^{\intercal} g_y(\vec{x},\vec{\theta})  \\ & - \sum_{y,y^{\prime}=2}^{M} q_{\vec{\theta}}(y|\vec{x}) q_{\vec{\theta}}(y^{\prime}|\vec{x}) \nabla_{\vec{x}} g_y(\vec{x},\vec{\theta}) \nabla_{\vec{x}} g_{y^{\prime}}^{\intercal}(\vec{x},\vec{\theta}). \nonumber
\end{align}
Let $\vec{v} \in \mathcal{X}$ be an arbitrary vector and define $\beta_y \doteq \nabla_{\vec{x}}^{\intercal} g_y(\vec{x},\vec{\theta}) \, \vec{v}$ for $y = [2:M]$. Notice that
\begin{align} \vec{G}(\vec{x}) \vec{v} & = \sum_{y=2}^{M} q_{\vec{\theta}}(y|\vec{x}) \beta_y \nabla_{\vec{x}} g_y(\vec{x},\vec{\theta}) \nonumber \\ & - \sum_{y=2}^{M} \left( \sum_{y'=2}^M  q_{\vec{\theta}}(y^{\prime}|\vec{x}) \beta_{y^{\prime}} \right) q_{\vec{\theta}}(y|\vec{x}) \nabla_{\vec{x}} g_y(\vec{x},\vec{\theta}). \end{align}
Therefore, the range of $\vec{G}(\vec{x})$ is a subset of the span of the set $\{ \nabla_{\vec{x}} g_2(\vec{x},\vec{\theta}), \ldots, \nabla_{\vec{x}} g_M(\vec{x},\vec{\theta}) \}$. Thus, it follows that $\text{rank}(\vec{G}(\vec{x})) \le M-1$, which implies that it is singular.

The singularity issue can be overcome by embedding $\mathcal{C}$ into the probability simplex $\mathcal{P}$ defined as follows: 
\begin{equation} 
\mathcal{P} = \Big\{ 
q : \mathcal{Y} \to [0,1]^M  : \sum_{y \in \mathcal{Y}} q(y) = 1 
\Big\}. \end{equation}
To proceed, we follow \cite{calin2014}[Section 2.8] and consider the following parameterization for any distribution $q \in \mathcal{P}$:
\begin{equation} q(y|\vec{z}) = \frac{z_y^2}{4}, \quad y \in\{ 1, \ldots, M\}. \end{equation}
We then consider the following statistical manifold:
\begin{equation} 
\mathcal{D} = \Big\{ q(\cdot|\vec{z}) : \|\vec{z}\|^2 = 4, z_y \ge 0,   \; \forall y \in \mathcal{Y} \Big\}. 
\end{equation}
Notice that the parameter vector $\vec{z}$ belongs to the positive portion of a sphere of radius 2 and centered at the origin in $\R^M$. As a consequence, the FIM follows by 
\begin{align} \label{eq:fim_simplex_reparam} \vec{G}(\vec{z}) & = \sum_{y\in\mathcal{Y}} \frac{z_y^2}{4} \left(\frac{2}{z_y} \vec{e}_y \right) \left(\frac{2}{z_y} \vec{e}_y^{\intercal} \right) = \vec{I}, \end{align}
where $\{\vec{e}_y\}$ are the canonical basis vectors in $\R^M$ and $\vec{I}$ is the identity matrix. From expression  \eqref{eq:fim_simplex_reparam} we can conclude that the Fisher metric is equal to the Euclidean metric. Since the parameter vector lies on a sphere, the \textsc{FRD} between the distributions $q = q(\cdot|\vec{z})$ and $q^{\prime} = q(\cdot|\vec{z}^{\prime})$ can be written as the radius of the sphere times the angle between the vectors $\vec{z}$ and $\vec{z}^{\prime}$. This leads to 
\begin{align} d_{R,\mathcal{D}}(q,q^{\prime}) & = 2 \arccos\left( \frac{\vec{z}^{\intercal} \vec{z}^{\prime}}{4}\right) \nonumber \\ & = 2 \arccos\left( \sum_{y\in\mathcal{Y}} \sqrt{q(y|\vec{z}) q(y|\vec{z}^{\prime})}\right). 
\end{align}
%
Finally, we can compute the \textsc{FRD} for distributions in $\mathcal{C}$ using: 
\begin{align} d_{R,\mathcal{C}}(q_{\vec{\theta}},q_{\vec{\theta}}^\prime) = 2 \arccos \left( \sum_{y \in \mathcal{Y}} \sqrt{q_{\vec{\theta}}(y|\vec{x} ) q_{\vec{\theta}}(y|\vec{x}^\prime)} \right). \end{align}
Notice that $0 \le d_{R,\mathcal{C}}(q_{\vec{\theta}},q_{\vec{\theta}}^\prime) \le \pi$, $\forall \ \vec{x},\vec{x}^\prime \in \mathcal{X}$, being zero if  $q_{\vec{\theta}}(\cdot|\vec{x}) = q_{\vec{\theta}}(\cdot|\vec{x}^\prime)$ and $\pi$ when $[q_{\vec{\theta}}(1|\vec{x}), \dots, q_{\vec{\theta}}(M|\vec{x})]$ and $[q_{\vec{\theta}}(1|\vec{x}^\prime), \dots, q_{\vec{\theta}}(M|\vec{x}^\prime)]$ are orthogonal vectors. 

\textbf{Proof the consistency between Theorem \ref{thm:binary_frd} and Theorem \ref{thm:multiclass_frd}}. This consists in showing the equivalence between the \textsc{FRD} for the binary case, given by Eq. \eqref{eq:fisher_rao_binary}, and the multiclass case, given by Eq. \eqref{eq:fisher_rao_multiclass}, for $M=2$. First, notice that the models \eqref{eq:binary_model} and \eqref{eq:standard_model} coincide if we consider the following correspondence: $h_{\vec{\theta}}(\vec{x}) = h_2(\vec{x},\vec{\theta}) - h_1(\vec{x},\vec{\theta})$, $y = 1 \leftrightarrow y = -1$ and $y = 2 \leftrightarrow y = 1$. Then, using standard trigonometric identities, we rewrite the \textsc{FRD} for the multiclass case:
\begin{align} \label{eq:rao_distance_multiclass2}
& d_{R,\mathcal{C}}(q_{\vec{\theta}},q_{\vec{\theta}}^\prime) = 2 \arccos \left( \sqrt{\frac{1}{1 + e^{h_{\vec{\theta}}(\vec{x})}}} \sqrt{\frac{1}{1 + e^{h_{\vec{\theta}}(\vec{x}^\prime)}}} \right. \nonumber \\ & \left. + \sqrt{\frac{1}{1 + e^{-h_{\vec{\theta}}(\vec{x})}}} \sqrt{\frac{1}{1 + e^{-h_{\vec{\theta}}(\vec{x}^\prime)}}}\right) \nonumber \\ & = 2 \arccos \left[ \cos(\arctan(e^{h_{\vec{\theta}}(\vec{x})/2})) \cos(\arctan(e^{h_{\vec{\theta}}(\vec{x}^\prime)/2})) \right. \nonumber \\ & \left. + \sin(\arctan(e^{h_{\vec{\theta}}(\vec{x})/2})) \sin(\arctan(e^{h_{\vec{\theta}}(\vec{x}^\prime)/2}))\right] \nonumber \\ & = 2 \arccos \left[ \cos \left( \arctan(e^{h_{\vec{\theta}}(\vec{x})/2}) - \arctan(e^{h_{\vec{\theta}}(\vec{x}^\prime)/2}) \right) \right] \nonumber \\ & = 2 \left| \arctan(e^{h_{\vec{\theta}}(\vec{x}^\prime)/2}) - \arctan(e^{h_{\vec{\theta}}(\vec{x})/2}) \right|, 
\end{align}
where the last step follows by $|\arctan(\alpha)| \le \pi/2$,  $\forall \alpha \in \R$, so the argument of the cosine function belongs to $[-\pi,\pi]$ and $\arccos(\cos(\alpha)) = |\alpha|$ for $|\alpha| \le \pi$, which completes the proof.

\subsection{Proof of Proposition \ref{prop-pareto}}
\label{AppendixF}


For completeness, we first present the derivation of the misclassification probabilities \eqref{eq:pe_nat_gaussian} and \eqref{eq:pe_adv_gaussian}: 
\begin{align}  P_e(\vec{\theta}) & = \P(f_{\vec{\theta}}(\vec{X}) \ne Y) = 
\Phi\left( - \frac{\theta^{\intercal} \vec{\mu}}{\sqrt{\theta^{\intercal} \vec{\Sigma} \theta}} \right), \\
P_e^{\prime}(\vec{\theta}) & = \P(f_{\vec{\theta}}(\vec{X}^{\prime*}) \ne Y) = 
\Phi\left( \frac{\varepsilon \| \vec{\theta}\|_2 - \theta^{\intercal} \vec{\mu}}{\sqrt{\theta^{\intercal} \vec{\Sigma} \theta}} \right). 
\end{align}
Notice that $\Phi$, i.e., the cumulative distribution function of the standard normal random variable, is a monotonic increasing function and $\varepsilon \| \vec{\theta}\|_2 / \sqrt{\theta^{\intercal} \vec{\Sigma} \theta} \ge 0$, so we have $P_e^{\prime}(\vec{\theta}) \ge P_e(\vec{\theta})$, as expected. Furthermore,  observe also that $\Phi$ is invertible so we can write $P_e^{\prime}(\vec{\theta})$ explicitly as a function of $P_e(\vec{\theta})$:
\begin{equation} \label{eq:pe_adv_funtion_pe_nat} P_e^{\prime}(\vec{\theta}) = \Phi \left(  \frac{\varepsilon \| \vec{\theta}\|_2}{\sqrt{\theta^{\intercal} \vec{\Sigma} \theta}} + \Phi^{-1}(P_e) \right). \end{equation}
We now proceed to the proof of the proposition. Notice that by the Rayleigh theorem \cite{horn2013matrix}[Theorem 4.2.2], we have that
\begin{equation} \lambda_{\text{min}}(\vec{\Sigma}) \| \vec{\theta} \|_2^2 \le \theta^{\intercal} \vec{\Sigma} \theta \le \lambda_{\text{max}}(\vec{\Sigma}) \| \vec{\theta} \|_2^2, \end{equation}
where $\lambda_{\text{min}}(\vec{\Sigma})$ and $\lambda_{\text{max}}(\vec{\Sigma})$ are the minimum and maximum eigenvalues of $\vec{\Sigma}$, respectively. Therefore, we can bound $\varepsilon \| \vec{\theta}\|_2 / \sqrt{\theta^{\intercal} \vec{\Sigma} \theta}$ as follows:
\begin{equation} \frac{\varepsilon}{\lambda_{\text{max}}^{1/2}(\vec{\Sigma})} \le \frac{\varepsilon \| \vec{\theta}\|_2}{\sqrt{\theta^{\intercal} \vec{\Sigma} \theta}} \le \frac{\varepsilon}{\lambda_{\text{min}}^{1/2}(\vec{\Sigma})}.  
\end{equation}
Using this fact together with the monotonicity of $\Phi$ in expression  \eqref{eq:pe_adv_funtion_pe_nat}, we obtain the desired result.





\section{Summary and Concluding Remarks}





We introduced  \textsc{Fire}, a new robustness regularizer-based method on the geodesic distance of softmax probabilities using concepts from information geometry. The main innovation is to employ Fisher-Rao Distance (\textsc{FRD}) to encourage invariant softmax probabilities for both natural and adversarial examples while maintaining high performances on natural samples. Our empirical results showed that \textsc{Fire} consistently enhances the robustness of neural networks using various architectures, settings, and datasets. Compared to the state-of-the-art methods for adversarial defenses, \textsc{Fire}  increases the Average Accuracy ({Avg. Acc.}). Besides, it succeeds in doing so with a 20\% reduction in terms of the training time.

Interestingly,  \textsc{FRD} has rich connections with Hellinger distance, the Kullback-Leibler divergence, and even other standard regularization terms. Moreover, as illustrated via our simple logistic regression and Gaussian model, the optimization based on \textsc{Fire} is well-behaved and gives all the desired Pareto-optimal points in the natural-adversarial region. This observation contrasts with the results of other state-of-the-art adversarial learning approaches. \textcolor{black}{Further theoretical explanation of this change in behaviour is worth exploring in a future work.}

\section*{Acknowledgment}

The authors wish to thank the Associate Editor and the 
reviewers for their
suggestions which significantly improved the manuscript.
This work was supported by the NSERC, and McGill University in the framework of the NSERC/Hydro-Quebec Industrial Research Chair in Interactive Information Infrastructure for the Power Grid (IRCPJ406021-14). This work was performed using
HPC resources from GENCI-IDRIS (Grant 2021-AD011012277 and AD011013109). This project has received funding from the European Union’s Horizon 2020 research and innovation program under the Marie Skłodowska-Curie grant agreement No 792464. 


\ifCLASSOPTIONcaptionsoff
\newpage
\fi

\bibliographystyle{IEEEtran}
\bibliography{IEEEabrv,refs}

\newpage
\appendices

\section{Comparison between the Fisher-Rao distance and the KL divergence on real data}\label{app:rao_vs_kl}
\begin{figure*}
    \centering
    \begin{subfigure}[b]{1\columnwidth}
		\centering
		\includegraphics[width=1\textwidth]{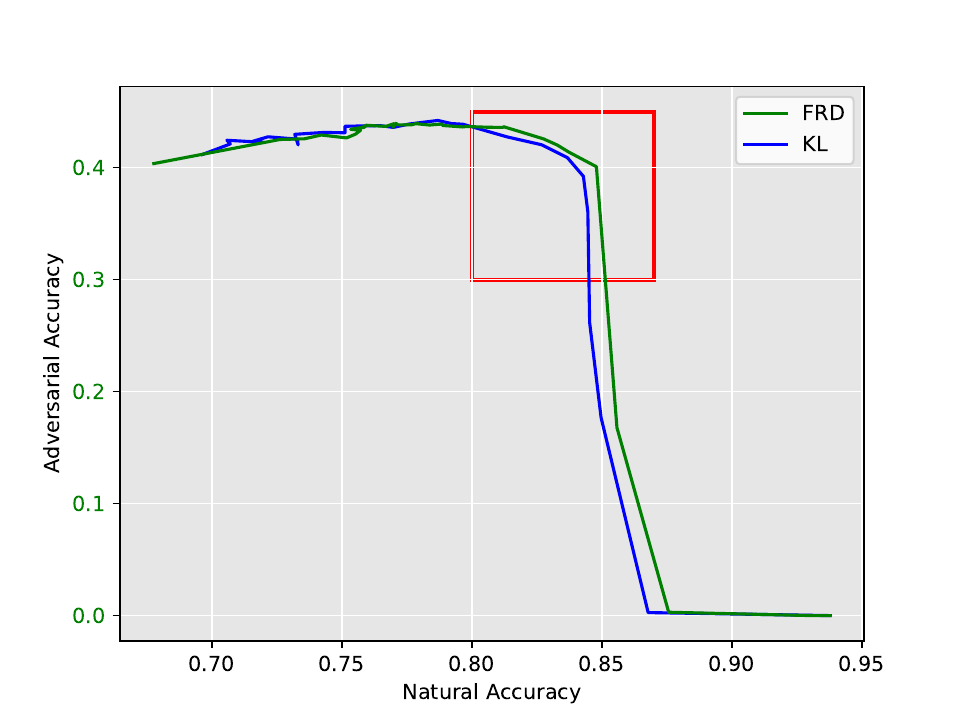}
	\end{subfigure}
    \begin{subfigure}[b]{1\columnwidth}
		\centering
    \includegraphics[width=1\textwidth]{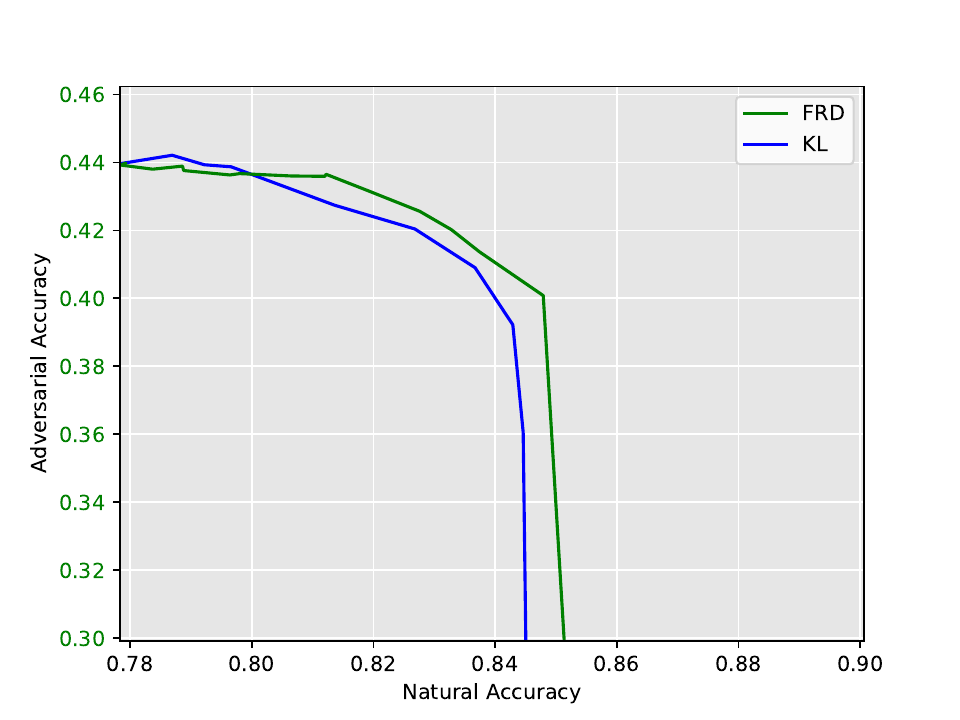}
    \end{subfigure}
    \caption{ Plots of all the possible points $(1-P_e(\vec{\theta}), 1-P_e^{\prime}(\vec{\theta}))$ for ResNet-18 model on CIFAR-10.}
    \label{fig:pareto_cifar}
\end{figure*}
In order to confirm on real data the difference between the Kullback-Leibler divergence and the Fisher-Rao distance, we reproduce the simulation presented on Fig. \ref{fig:nat_vs_adv_acc_gaussian} based on the CIFAR-10 dataset. We trained multiple classifier using both TRADES  and  FIRE methods, varying $\lambda\in [0, 50]$.  We use the dataset CIFAR-10 with a ResNet-18 for the model, and train the models for $100$ epochs. The optimizer is the Stochastic Gradient Descent (SGD) with a learning rate of 0.01, with a decay of 0.1 at epoch 75, 90 and 100. We also use a weight decay of $5.10^{-4}$, a momentum of 0.9. The results are averaged over 2 tries.


We present the results on Fig. \ref{fig:pareto_cifar}. On the left figure, we observe   the entire curve, and on the right we zoomed on the zone of interest. We can see that Fisher-Rao distance presents a better trade-off between natural and adversairal accuracies than the Kullback-Leibler divergence on real data, confirming the improvements presented on Fig. \ref{fig:nat_vs_adv_acc_gaussian}. For natural accuracies below 80\%, the Fisher-Rao distance and the Kullback-Leibler divergence seem to behave quite similarly. For natural accuracies above 80\%,  which is actually the zone of interest, the improvement caused by the use of the Fisher-Rao distance seem to be quite consistent among all the training points. At fixed adversarial accuracies, the Fisher-Rao distance can increase the results by up to 1\% of natural accuracies. Note that, in this simulation the Kullback-Leibler divergence can achieve a better adversarial accuracy that the Fisher-Rao distance (44.21\% compared to 43.57\%), but with a cost of sighly more than 2.5\% for natural accuracies (78.69\% compared to 81.23\%). The Fisher-Rao divergence therefore achieves a better trade-off between natural and adversarial accuracies than the Kullback-Leibler divergence. 

\begin{figure*}
	\centering
	\begin{subfigure}[b]{0.65\columnwidth}
		\centering
		\includegraphics[width=1.15\columnwidth]{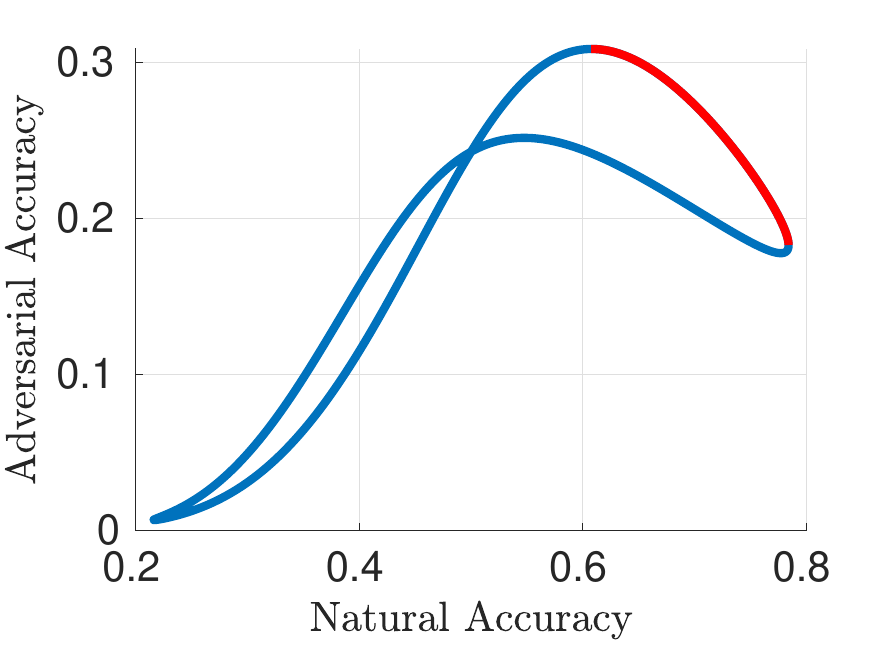}
		\caption{ Pareto-optimal points}
		\label{fig:nat_vs_adv_acc_gaussian_pareto_2}
	\end{subfigure}
	\hfill
\begin{subfigure}[b]{0.65\columnwidth}
		\centering
		\includegraphics[width=1.15\columnwidth]{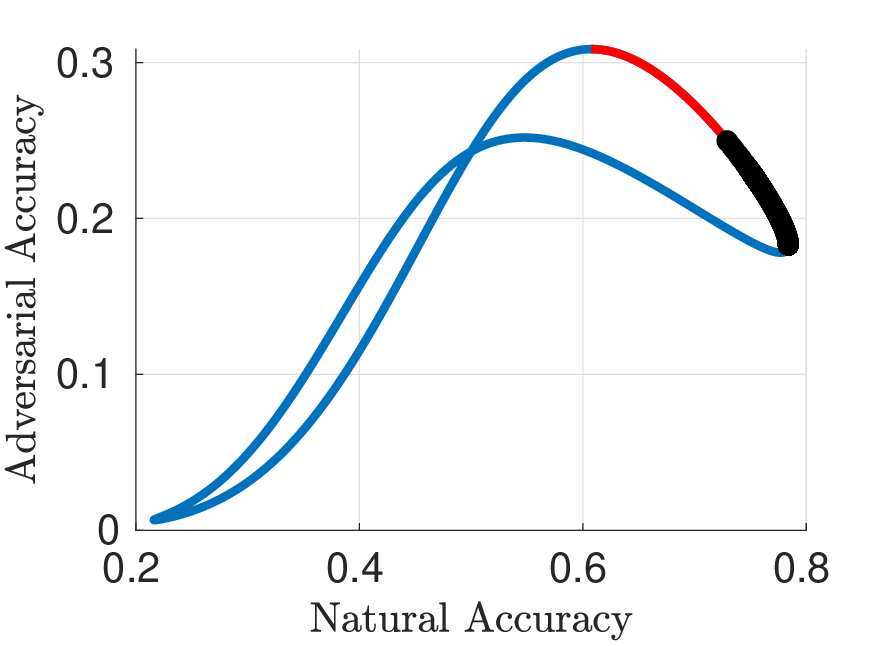}
		\caption{ Hellinger}
		\label{fig:nat_vs_adv_acc_gaussian_hel_2}
	\end{subfigure}
	\hfill
	\begin{subfigure}[b]{0.65\columnwidth}
		\centering
		\includegraphics[width=1.15\columnwidth]{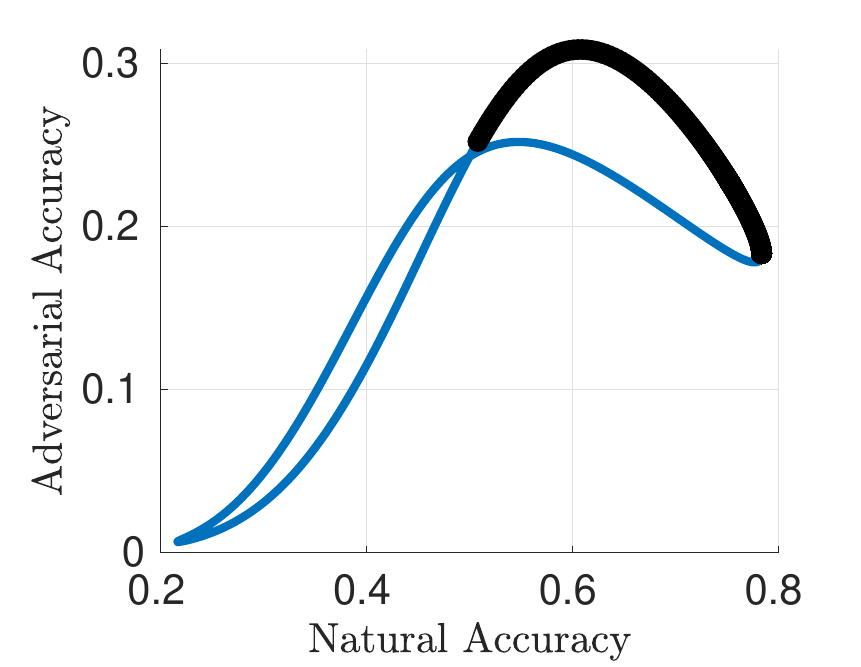}
		\caption{ FIRE}
		\label{fig:nat_vs_adv_acc_gaussian_fire_2}
	\end{subfigure}
	\hfill
	
	\caption{ Plots of all the possible points $(1-P_e(\vec{\theta}), 1-P_e^{\prime}(\vec{\theta}))$ for the Gaussian model with $\varepsilon = 0.1$, $\vec{\mu} = [-0.0218; 0.0425]$ and $\vec{\Sigma} = [ 0.0212, 0.0036; 0.0036, 0.0042]$ shown in blue. In red, we show the Pareto-optimal points (Fig. \ref{fig:nat_vs_adv_acc_gaussian_pareto_2}). In black, we show the solutions obtained by minimizing the risk  $L_{\textrm{Hel}}(\vec{\theta})$ in \eqref{eq:regularized_risk_hel} (Fig. \ref{fig:nat_vs_adv_acc_gaussian_hel_2}), the risk  $L_{\textrm{FIRE}}(\vec{\theta})$ in  \eqref{eq:regularized_risk_fire} (Fig. \ref{fig:nat_vs_adv_acc_gaussian_fire_2}).
	}
	\label{fig:nat_vs_adv_acc_gaussian_2}
\end{figure*}

\begin{table*}[t]
	\tiny
	\centering
	\caption{{ Comparison between Hellinger and Fisher-Rao based regularizer  under white-box $l_{\infty}$ threat model.}}
	\resizebox{0.84\textwidth}{!}{
		\begin{tabular}{c||c|c|c||c|c||c||c}
			\toprule
			
			Defense & Dataset &$\varepsilon$& Structure & Natural  & AutoAttack & { Avg. Acc.} & RunTime\\
			\midrule
			\midrule
			
			\rowcolor{Gray} { Hellinger} & & & { CNN} & {\textbf{ 99.31 $\pm$ 0.03}} & { 94.03 $\pm$ 0.24 } & {  96.67 $\pm$ 0.07 } & {  2h06}\\
			\rowcolor{Gray}{  FIRE} & { \multirow{-2}{*}{MNIST}}&{ \multirow{-2}{*}{$0.3$}} & { CNN} & { 99.22 $\pm$ 0.02} & { \textbf{94.44 $\pm$ 0.14}} & { \textbf{96.83} $\pm$ 0.10} & { \textbf{2h06}} \\
			\midrule
			
			{  Hellinger} &  & & { WRN-34-10} & { 85.96 $\pm$ 0.28} &  { 50.52 $\pm$ 0.31} & { 68.24 $\pm$ 0.15} & { 11h02} \\
			\rowcolor{Gray} { FIRE} & { \multirow{-2}{*}{CIFAR-10}} & { \multirow{-2}{*}{$8/255$}}& { WRN-34-10} & \textbf{{ 85.98 $\pm$ 0.09}}  & \textbf{{ 51.45 $\pm$ 0.32}} & \textbf{{ 68.72 $\pm$ 0.22}} & { \textbf{11h00}}\\
			\midrule
			
			{ Hellinger} & & & { WRN-34-10} & {  60.79 $\pm$ 0.88 } & {  25.58 $\pm$ 0.27 }  & {  43.18 $\pm$ 0.54} & {  11h12 } \\
			\rowcolor{Gray}  { FIRE} & { \multirow{-2}{*}{CIFAR-100}} & { \multirow{-2}{*}{$8/255$}}& { WRN-34-10} & \textbf{{  61.03 $\pm$ 0.21}} &  \textbf{{ 26.42 $\pm$ 0.21}} & \textbf{{  43.73 $\pm$ 0.12}}  & { \textbf{11h10}} \\
			
			\bottomrule
	\end{tabular}}
	\label{tab:comparison_hellinger}
\end{table*}

\section{Comparison between Fisher-Rao and Hellinger distances}

\subsection{Theoretical relation} \label{sec:frd_vs_hel}

The \emph{Hellinger} distance between two distributions $q$ and $q^\prime$  on $\mathcal{Y}$ is defined as follows: 
\begin{equation} \H(q,q^\prime) \doteq \sqrt{2} \left( 1 - \sum\limits_{y \in \mathcal{Y}} \sqrt{q(y) q(y)^\prime}\right)^{1/2}. \label{eq:hellinger_definition} \end{equation}
Using \eqref{eq:fisher_rao_multiclass}, we readily obtain the following relation between the Hellinger distance and the Fisher-Rao distance.

\begin{theorem}[Relation between \textsc{FRD} and Hellinger distance] The \textsc{FRD} between soft-predictions $q_{\vec{\theta}} = q_{\vec{\theta}}(\cdot|\vec{x})$ and $q_{\vec{\theta}}^\prime = q_{\vec{\theta}}(\cdot|\vec{x}^\prime)$, given by \eqref{eq:fisher_rao_multiclass} is related to the Hellinger distance through the relation
\begin{equation} \label{eq:hellinger_vs_rao} \H^{2}(q_{\vec{\theta}},q_{\vec{\theta}}^\prime) = 2 \left[ 1 - \cos \left(\frac{d_{R,\mathcal{C}}(q_{\vec{\theta}},q_{\vec{\theta}}^\prime)}{2}\right) \right]. \end{equation}
Since $0 \le d_{R,\mathcal{C}}(q_{\vec{\theta}},q_{\vec{\theta}}^\prime) \le \pi$, it is clear that $\H^{2}(q_{\vec{\theta}},q_{\vec{\theta}}^\prime)$ is a monotonically increasing function of $d_{R,\mathcal{C}}(q_{\vec{\theta}},q_{\vec{\theta}}^\prime)$.
\end{theorem}

We conclude from this result that the FRD and the Hellinger distance are \emph{theoretically equivalent} regularization mechanisms. However, it is clear that the empirical optimization of these distances may be different. This is further explored and confirmed in the following.

\subsection{Experimental comparison}
Since the Hellinger distance and the Fisher-Rao distance are theoretically equivalent metrics (see Section \ref{sec:frd_vs_hel}), we investigate the empirical difference between those two distances. First, we perform the same simulations as those presented in Fig \ref{fig:nat_vs_adv_acc_gaussian} but using the Hellinger distance as a robust regulatizer. Then, we  perform  comparison between Hellinger and FRD on real datasets (similar to the simulations given in Section \ref{sec:comparison_kl}).
\subsubsection{Accuracy-Robustness trade-offs in the Gaussian case}
As we defined the FIRE risk, it is possible to define the Hellinger risk as :  
\begin{align} 
L_{\textrm{\textsc{Hel}}}(\vec{\theta})  \doteq & \, \E_{p(\vec{x},y)}\Big[ \max_{\vec{x}^{\prime} : \|\vec{x}^{\prime} - \vec{x} \|_{p}\leq \varepsilon} -\log q_{\vec{\theta}}(y|\vec{x}) \nonumber \\ & + \lambda \;  H^{2}(q_{\theta}(\cdot|\vec{x}), q_{\theta}(\cdot|\vec{x}^{\prime}))) \Big], \label{eq:regularized_risk_hel}
\end{align}
where $H(q_{\theta},q_{\theta}^{\prime}) $ is defined in equation \eqref{eq:hellinger_definition}. 
 In Fig. \ref{fig:nat_vs_adv_acc_gaussian_kl}, we present the solution of the (local) Empirical Risk Minimization (ERM) for the Hellinger risk function as defined in \eqref{eq:regularized_risk_hel} for different values of $\lambda$. As can be observed, the curve obtained for all pairs of $(1-P_e(\vec{\theta}), 1-P_e^{\prime}(\vec{\theta}))$  covers about half of the Pareto-optimal points while the curve of all solutions for $(1-P_e(\vec{\theta}), 1-P_e^{\prime}(\vec{\theta}))$ corresponding  to FIRE risk (see \eqref{eq:regularized_risk_fire}) covers all the Pareto-optimal points. 
This simulation shows that even if the Hellinger distance and the Fisher-Rao distance are theoretically equivalent, their dynamics in training are actually quite different. In addition, FRD seems to be better suited for training in the Gaussian case. 

{ 
\subsubsection{Comparison based on real data}



We now study the empirical difference between those two distances on real datasets. We therefore perform the same simulations as those provided  in subsection \ref{sec:comparison_kl} using the squared Hellinger distance between the natural and the adversarial predictions as the robustness regularizer. The results are summarized in Table \ref{tab:comparison_hellinger}. Even though using the Hellinger distance as a robust regularizer performs better than using the Kullback-Leibler divergence, it still performs worse that the Fisher-Rao distance. In the case of real data, FRD appears to  be better suited for training than the Hellinger distance.

In conclusion, FRD provides a better regularization objective than the Hellinger distance.  }

\section{Proof of Proposition \ref{thm:fisher_rao_bound_binary} and Theorem \ref{thm:relation_kl_and_frd}}

\subsection{Proof of Proposition \ref{thm:fisher_rao_bound_binary}}
\label{AppendixC}

Notice that the \textsc{FRD} in  \eqref{eq:fisher_rao_binary} can be written as follows:
\begin{align} d_{R,{\mathcal{C}}}(q_{\vec{\theta}}, q_{\vec{\theta}}^\prime) = 2 \Big|f(h_{\vec{\theta}}(\vec{x}^\prime)) - f(h_{\vec{\theta}}(\vec{x}))\Big|, \end{align}
where we have defined the function $f:\R \to \R$ as $f(z) \doteq \arctan(e^{z/2})$. Notice that $f$ is a smooth function. Using the mean value theorem, we have
\begin{equation} d_{R,{\mathcal{C}}}(q_{\vec{\theta}}, q_{\vec{\theta}}^\prime) = 2 |f^{\prime}(c)| \cdot  \big |h_{\vec{\theta}}(\vec{x}^\prime) - h_{\vec{\theta}}(\vec{x})\big|, \end{equation}
where $f^{\prime}(z) = e^{z/2} / (2 e^z + 2)$ is the derivative of $f$ and $c$ is a point in the (open) interval with endpoints $h_{\vec{\theta}}(\vec{x})$ and $h_{\vec{\theta}}(\vec{x}^\prime)$. Notice that $0 \le f^{\prime}(z) \le 1/4$ for any $z$. Then, we have
\begin{align} d_{R,{\mathcal{C}}}(q_{\vec{\theta}}, q_{\vec{\theta}}^\prime) & \le \underset{c}{\text{sup}} \; 2 |f^{\prime}(c)|\cdot  \big|h_{\vec{\theta}}(\vec{x}^\prime) - h_{\vec{\theta}}(\vec{x})\big| \\ & = \frac{1}{2} \big|h_{\vec{\theta}}(\vec{x}^\prime) - h_{\vec{\theta}}(\vec{x})\big|. \end{align}

\subsection{Proof of Theorem \ref{thm:relation_kl_and_frd}}
\label{AppendixE}

Consider first the \emph{Hellinger} distance between two distributions $q$ and $q^\prime$ over $\mathcal{Y}$, defined as follows: 
\begin{equation*} \H(q,q^\prime) \doteq \sqrt{2} \left( 1 - \sum\limits_{y \in \mathcal{Y}} \sqrt{q(y) q(y)^\prime}\right)^{1/2}. \end{equation*}
Using \eqref{eq:fisher_rao_multiclass}, we readily obtain the following relation:
\begin{equation} \label{eq:hellinger_and_rao} \H(q,q^\prime) = \sqrt{2} \left[ 1 - \cos \left(\frac{d_{R,\mathcal{C}}(q,q^\prime)}{2}\right) \right]^{1/2}. \end{equation}
We now use the following inequality relating the Hellinger distance and the KL divergence \cite[Lemma 2.4]{tsybakov2008introduction}:
\begin{equation} \H^2(q,q^\prime) \le \KL(q,q^\prime). \end{equation}
Using the relation \eqref{eq:hellinger_and_rao}, we obtain the desired bound \eqref{eq:rao_bound_kl}. The second-order approximation \eqref{eq:second_order_rao_kl} follows directly from \cite{calin2014}[Theorem~4.4.5].


\end{document}